\documentclass{ieeetj}
\usepackage{cite}
\usepackage{amsmath,amssymb,amsfonts}
\usepackage{algorithmic}
\usepackage{graphicx,color}
\usepackage{textcomp}
\usepackage{xcolor}
\usepackage{hyperref}
\usepackage[noabbrev,capitalise]{cleveref}
\usepackage{booktabs}
\usepackage{helper_glossaries}
\usepackage{units}
\hypersetup{hidelinks=true}
\usepackage{algorithm,algorithmic}
\def\BibTeX{{\rm B\kern-.05em{\sc i\kern-.025em b}\kern-.08em
    T\kern-.1667em\lower.7ex\hbox{E}\kern-.125emX}}
\AtBeginDocument{\definecolor{tmlcncolor}{cmyk}{0.93,0.59,0.15,0.02}\definecolor{NavyBlue}{RGB}{0,86,125}}

\usepackage[caption=false,font=footnotesize]{subfig}
\usepackage[numbers]{natbib}
\newcommand{\norm}[1]{\left\lVert#1\right\rVert}

\newcommand{\pos}{\boldsymbol{p}}
\newcommand{\quat}{\boldsymbol{q}}
\newcommand{\rotmat}{{R}}
\newcommand{\eulang}{\boldsymbol{\mu}}
\newcommand{\vel}{\boldsymbol{v}}
\newcommand{\angvel}{\boldsymbol{\omega}}
\newcommand{\veltwist}{\boldsymbol{\nu}}
\newcommand{\pose}{\boldsymbol{\eta}}

\newcommand{\inertia}{M}
\newcommand{\coriolis}{C}
\newcommand{\drag}{D}
\newcommand{\gravity}{{g}}

\newcommand{\wrench}{\mathbf{w}}

\newcommand{\actforce}{\boldsymbol{f}}
\newcommand{\distforce}{\actforce_d}

\newcommand{\actinput}{\boldsymbol{u}}

\newcommand{\samplingtime}{\ensuremath{\Delta_t}}

\newcommand{\residual}{\boldsymbol{\delta}}

\newcommand{\gtpose}{\pose^{\text{gt}}}
\newcommand{\gtveltwist}{\veltwist^{\text{gt}}}

\newcommand{\simveltwist}{\veltwist^{\text{sim}}}

\newcommand{\dataset}{\ensuremath{\mathcal{D}_K}}

\newcommand{\fsam}{\ensuremath{f_{\text{s}}}}

\newcommand{\vect}[1]{\boldsymbol{#1}}

\def\OJlogo{\vspace{-4pt}$<$Society logo(s) and publication title will appear here.$>$}
\def\seclogo{\vspace{10pt}$<$Society logo(s) and publication title will appear here.$>$}

\def\authorrefmark#1{\ensuremath{^{\textbf{#1}}}}

\begin{document}
\receiveddate{XX Month, XXXX}
\reviseddate{XX Month, XXXX}
\accepteddate{XX Month, XXXX}
\publisheddate{XX Month, XXXX}
\currentdate{XX Month, XXXX}
\doiinfo{XXXX.2022.1234567}

\markboth{}{Author {et al.}}

\title{Marinarium: A Modular Experimental Facility for Reproducible Maritime and Space-Analog Field Robotics}

\author{Ignacio Torroba\authorrefmark{1}, David Dörner\authorrefmark{1}, Victor Nan Fernandez-Ayala\authorrefmark{2}, Mart Kartasev\authorrefmark{3}, Joris Verhagen\authorrefmark{3}, Elias Krantz\authorrefmark{1},
Gregorio Marchesini\authorrefmark{2}, Carl Ljung\authorrefmark{1}, Pedro Roque\authorrefmark{4}, Chelsea Sidrane\authorrefmark{3}, Linda Van der Spaa\authorrefmark{2}, Nicola De Carli\authorrefmark{2}, Petter Ögren\authorrefmark{3}, Christer Fuglesang\authorrefmark{1}, 
Jana Tumova\authorrefmark{3}, Dimos V. Dimarogonas\authorrefmark{2} and Ivan Stenius\authorrefmark{1}}
\affil{Division of Aerospace, Moveability and Naval Architecture, Royal Institute of Technology, 10044 Stockholm, Sweden}
\affil{Department of Decision and Control Systems, Royal Institute of Technology, 10044 Stockholm, Sweden}
\affil{Department of Robotics, Perception and Learning, Royal Institute of Technology, 10044 Stockholm, Sweden}
\affil{Division of Engineering and Applied Science, California Institute of Technology, Pasadena, CA 91125, USA}
\corresp{Corresponding author: Ignacio Torroba (email: torroba@kth.se).}
\authornote{This work was partially supported by Digital Futures, Vinnova and FMV through the SHARCEX grant and by the Wallenberg AI, Autonomous Systems and Software Program (WASP) funded by the Knut and Alice Wallenberg Foundation.}

\begin{abstract}
Field robotics research in maritime and space domains is constrained by a persistent gap between low-cost, low-fidelity simulation and costly offshore experimentation. Instrumented water tanks partially bridge this gap but often provide limited sensing, restricted experimental capabilities, and weak integration with simulation tools. To address these limitations, we present Marinarium, a modular, standalone experimental facility that provides a cost-effective intermediate testbed between simulation and field deployment. Marinarium combines a fully instrumented underwater and aerial operational volume with motion capture (MoCap), a retractable roof enabling both sheltered and open-air operation, a digital twin implemented in SMaRCSim, and direct integration with a planar space robotics laboratory, enabling both maritime and underwater space-analog experimentation. We present the design rationale of the facility and validate its capabilities through four representative studies in field robotics: (i) data-driven system identification of underwater vehicle dynamics; (ii) heterogeneous multi-domain robotic rendezvous; (iii) sim-to-real transfer for underwater robotics using learned dynamics residuals; and (iv) cross-domain validation of spacecraft autonomy using underwater surrogates. Together, these studies demonstrate that Marinarium enables reproducible, instrumented experimentation across multiple field robotics challenges that would otherwise require costly offshore deployments or be impractical to investigate using simulation alone.

\end{abstract}



\maketitle

\section{INTRODUCTION}
\IEEEPARstart{R}{esearch} and development of maritime robotics has traditionally lagged behind its ground and aerial counterparts \cite{zereik2018challenges}. The reasons for this gap are manifold: i) the complexity and risks associated with deploying and recovering robots offshore, which makes testing maritime technology intrinsically more costly and inaccessible, ii) the limited observability and weather-dependent repeatability of experiments carried out in open water, which hinders a rigorous evaluation and benchmarking of results, and iii) the reduced ability to compile large-scale datasets \gls{uw} with accurate ground truth, a foundational block of machine learning \cite{aubard2025sonar}. 

\begin{figure*}[t] 
    \centering
    \includegraphics[width=0.99\linewidth]{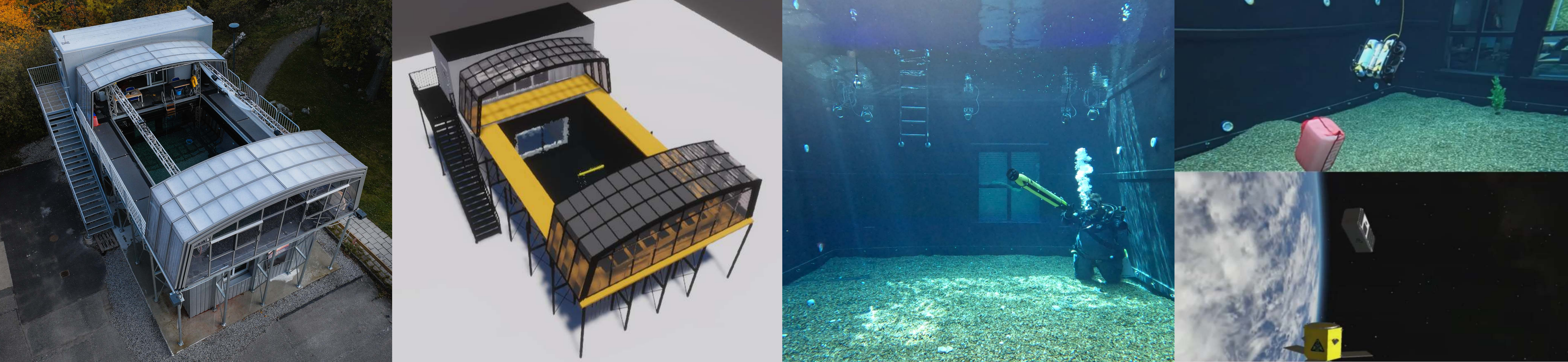}
    \caption{Left to right: i) aerial view of the Marinarium facility, ii) the facility digital twin in the SMaRCSim, iii) example of diver-AUV interaction experiment in the water basin and iv) equivalent inspection tasks executed by a BlueROV2 in the tank and a simulated CubeSat in Basilisk \cite{kenneally2020basilisk}.}
    \label{fig:carl}
\end{figure*}

To bypass these limitations, researchers in marine robotics often resort to simulators as widely accessible surrogates of the marine domain. The recent boom in open-source \gls{uw} simulators is a sign of this trend \cite{aldhaheri2025underwater, kartavsev2025smarcsim, manhaes2016uuv}. However, the complexity of accurately modeling hydrodynamic interactions, acoustic and electromagnetic propagation, and open water conditions results in large differences remaining between the simulation and the real open water domain. These differences, known as the simulation to real (Sim2Real) gap \cite{hofer2021sim2real}, severely limit the fidelity of this type of simulators and thus hinder the deployment of systems developed in simulation into the real world, resulting in underwater robotics research often remaining in the simulation realm \cite{chang2025learning, lidtke.2024}.

Water tank facilities are often used as an intermediate development step between simulation and deployment in the real maritime environment \cite{lindsay2022collaboration, ferreira2024remote}. They can serve as a proxy of the offshore domain with controllable and observable conditions, reducing testing risks and enabling repeatability of the experiments.
There is a large body of work containing instances of water tank facilities used as testbeds for experimentation in guidance, navigation and control of underwater vehicles \cite{kinsey2003new, xanthidis2020navigation, xu2022integrated}, multi-domain collaboration with heterogeneous fleets \cite{lindsay2022collaboration} or deployment of control policies trained in simulation \cite{cai2025learning, li2021deep, lin2026intelligent}.

Analogously, water tank facilities can also serve as accessible test environments for space robotics~\cite{carignan_reaction_stabilization, koch2020underwater}. Neutral buoyancy conditions underwater provide a useful approximation of microgravity, allowing experimentation and testing under reduced weight effects and continuous, low-damping motion~\cite{marciacq2009crew, neufeld2015practicing}. 
However, existing water tank facilities often lack the instrumentation and simulation integration needed to pursue systematic research in i) data-driven system identification for underwater vehicles, ii) multi-domain heterogeneous collaborative robotics, iii) Sim2Real bridging for underwater simulators, and iv) spacecraft navigation validation via underwater surrogates.

We address this gap by developing the Marinarium, a newly designed underwater robotics research facility at \gls{kth}. The Marinarium, shown in \cref{fig:carl}, is a state-of-the-art yet low-cost, self-contained, and modular water tank facility.
The Marinarium consists of a \unit[9$\times$5$\times$3]{m} water basin above the ground with a \unit[9$\times$5$\times$2]{m} drone space above the water's surface and a directly connected control room. The resulting operational volume is housed by a fully retractable ceiling to provide real weather conditions during surface and air operations if desired, and it is fully monitored both above and underwater via two \glspl{mocap}. 
These capabilities position the Marinarium as an intermediate field-robotics testbed: more controlled and observable than open water, yet substantially closer to offshore operating 
conditions than desktop simulation.

We argue that the following four contributions are key to the novelty and utility of this new infrastructure design to the field robotics research community:
\begin{itemize}
    \item A novel modular design for a cost-effective, standalone underwater research facility, enabling affordable, long-term experimentation and data collection. 
    \item A combined air and underwater operating volume, monitored with two \gls{mocap} systems and with access to the open air through a retractable dome, creating the ideal arena for inter-domain experimentation in air, surface and underwater robotics.
    \item A digital twin of the facility in the SMaRCSim \gls{uw} simulator \cite{kartavsev2025smarcsim}, enabling paired real-to-sim and Sim2Real studies on underwater perception and actuation.
    \item A tight integration with the \gls{kth} space robotics laboratory \cite{roque2025towards}, resulting in a well-suited testbed to study and exploit the commonalities between robotic platforms in space and underwater conditions. 
\end{itemize}

In the remainder of this paper, we present the current state of the art facilities for underwater robotics research, identify their limitations and use them to motivate our design. We then showcase 
how the resulting design choices help bring realistic maritime robotics closer to the research and academic community by presenting results in four open research areas enabled by them: 
data-driven \gls{sysid} for underwater vehicles, heterogeneous, multi-domain maritime robotics, underwater Sim2Real bridging, and space robotics validation via underwater surrogates.

\section{RELATED WORK}
In this section, we present an overview of research facilities similar to the Marinarium around the globe. Subsequently, we introduced related work in the four different research areas in which tank-based experiments have been executed.

\subsection{EXISTING MARINE ROBOTICS RESEARCH FACILITIES}
\begin{table*}
\centering
\caption{Similar underwater robotics research facilities worldwide and their main characteristics regarding location, volume, control of water conditions, motion tracking capabilities underwater (UW), additional connected \gls{aw} capabilities, suitability for divers --where `-' indicates a potential but not-known experiments involving divers--, and research focus. The following additional abbreviations are employed: \gls{rgb}, \gls{uwb}, \gls{usv}, \gls{uav}.}
\label{tab:uwf} 
\begin{tabular}{p{17mm}lcp{16mm}p{10mm}p{15mm}cp{37mm}}
\toprule
Institution & Location & L$\times$W$\times$D or $\varnothing\times$D (\unit[]{m}) & Control & \gls{uw} & \gls{aw} & Divers & Type of Research \\ 
\midrule
SINTEF~\cite{sintef-ocean-lab}
    & Norway  & $80\times50\times10$ & Wave, Wind, Current & \gls{mocap}, \gls{rgb} & \gls{mocap}, \gls{rgb} & - & \gls{uw}/Surface Structures, \glspl{auv} \\
TCOMS~\cite{TCOMS_RD}
    & Singapore & $60\times48\times12$ & Wave, Current & - & \gls{mocap} & no & \gls{uw} Structures, Robotics \\
~~~deep pit &  & $\varnothing 10\times50$ &  & - & - & no \\ 
Ifremer~\cite{brest} & France & $50\times12.5\times20$ & Wave, Wind & \gls{mocap}, \gls{rgb} & \gls{mocap}, \gls{rgb} & yes & Hydrodynamics, \gls{uw} Structures, Robotics incl. Diver Collab.\\
Aquatron, Pool Tank~\cite{aquatron_pool_tank}
    & Canada & $\varnothing 15.24\times3.91$ & Temperature, Current & - & Drone space & - & Marine Life, Multi-Domain, Robotics, Sensors~\cite{lindsay2022collaboration} \\
~~~Tower Tank & & $\varnothing 3.66 \times 10.46$ & Current & - & - & - & Pressure Activated Instruments\\
~~~Pools 1,2,3 & & $9.1 \times 7.3 \times 4$ & Current & - & - & - & \glspl{auv}, \glspl{usv} \\

CalTech~\cite{gunnarson2024fish}
    & USA & $4.8\times1.8\times1.5$ & Current & - & - & no & \gls{uw} Robotics \\
UMich~\cite{umich}
    & USA & $5.18\times3.05\times1.98$ &  & \gls{mocap} & \gls{mocap} & & \glspl{auv}, \glspl{usv}, \gls{uw} Robotics \\
FRoSt~\cite{FRoStLab_Tank}
    & USA & $9.14\times6.10\times1.83$ &  & - & -   & - & \glspl{auv} \\
JHU~\cite{kinsey2003new}
    & USA & $\varnothing 7.49\times3.96$ &  & -     & -    & - & \glspl{auv}, Robotics   \\
NBRF~\cite{akin_robotic_capabilities,UMD_NBRF}
    & USA & $\varnothing 15.24\times7.62$ &  &  \gls{mocap}, \gls{rgb}  & -  & yes & Space/\gls{uw} Robotics \\
NBL~\cite{jairala2012eva}
    & USA & $61.5\times31.1\times12.2$  & Temperature, Wind, Wave  & \gls{mocap} &  \gls{mocap}  & yes & Astronaut Training/Medical Research \\
EAC~\cite{marciacq2009crew}
    & Germany & $22\times17\times10$        &   & -   & -  & yes & Astronaut Training \\
DFKI~\cite{DFKI_MaritimeInfrastructure}
    & Germany   & $23\times19\times8$       &   & \gls{mocap}   &  \gls{rgb}  & yes & Space/\gls{uw} Robotics   \\
~~~glass tank & & $5\times4\times2.2$       &   & -    & -  & no & \gls{uw} Robotics \\
~~~black tank & & $3.4\times2.6\times2.2$ & Turbidity & - & - & no & \gls{uw} Robotics \\
AST~\cite{Fraunhofer_IOSB-AST}
    & Germany & $12\times8\times3$ & & - & - & - & \gls{uw} Robotics \\
CIRS~\cite{CIRS}
    & Spain & $16.5\times9.5\times5$ &      & \gls{rgb}   & -    & -  & \gls{uw} Robotics, \gls{uw} Vision \\
CIRTESU~\cite{monros2025surrogate}
    & Spain & $12\times8\times5$ & Current, Wave & -  & - & - & \gls{uw} Robotics, Vision, Wireless Comm., Teleop., Fluids \\
ISEP~\cite{INESC-TEC_CRAS_lab}
    & Portugal & $10\times6\times5$  &      & \gls{rgb}   & -   & - & \glspl{usv}, \gls{uw} Robotics \\
LABUST~\cite{ferreira2024remote} 
    & Croatia & $7.8\times4.1\times3$    &     & \gls{rgb}  & \gls{rgb}, \gls{uwb}, space for micro-\gls{uav}   & - & \gls{uw} Robotics \\
AAU~\cite{AAU_UnderwaterTechnology} 
    & Denmark & $7.5\times3.6\times1.5$ & & \gls{mocap} & - & - & \glspl{usv}, \gls{uw} Robotics \\
    ~~~black tank & & $1.2\times1.2\times2.3$ & Acoustic damping & - & - & no & Perception \\
ASTA~\cite{asta-dtu}
    & Denmark  & $6.5\times3.5\times3$ & Current &  -    & \gls{mocap}, \unit[16$\times$12$\times$8]{m} Drone cage & no & Multi-Domain Collab. Robotics\\
\bottomrule
\end{tabular}
\end{table*}

Based on their environmental scope, existing maritime robotics research facilities range from large natural outdoor areas for field testing to small indoor, fully-monitored basins with controlled conditions. 
On the one side of this range, facilities such as the MBARI MARS system~\cite{mcnutt2003mars}, the Fraunhofer \gls{dol} in the Baltic Sea~\cite{umlauft2024digital}, and the NTNU and SINTEF Fjordlab~\cite{ludvigsen2025fjordlab} provide sensors for environmental monitoring and robotics development. The \gls{dol} and Fjordlab complement stationary platforms with instrumented surface and underwater vehicles.
However, while extremely valuable, such facilities are very scarce due to their large scale and operational complexity and therefore fail to directly contribute to the democratization of underwater robotics research. 
On the other side of the spectrum, smaller indoor water tanks such as the Marinarium are more widely used in academic research due to their simpler setup and affordability. In \cref{tab:uwf}, we compiled a list of representative facilities and highlighted, for comparison, their main characteristics in terms of size and location, existing equipment and intended use. 

Regarding the size of the facility, the water basins housed by SINTEF Ocean~\cite{sintef-ocean-lab}, TCOMS~\cite{TCOMS_RD}, Ifremer~\cite{brest} and the Neutral Buoyancy Laboratory (NBL)~\cite{jairala2012eva} at NASA's \gls{jsc} are among the largest in the world. 
On the scale of the Marinarium, North American universities house \gls{uw} tanks at Caltech~\cite{gunnarson2024fish}, Dalhousie University~\cite{aquatron_pool_tank}, University of Michigan~\cite{umich}, Brigham Young University~\cite{FRoStLab_Tank}, and John Hopkins University~\cite{kinsey2003new}.
In mainland Europe, similar infrastructures for \gls{uw} robotics testing are found at the Fraunhofer Institute for Applied Systems Engineering~\cite{Fraunhofer_IOSB-AST} in Germany; CIRS~\cite{CIRS} at the University of Girona and CIRTESU~\cite{monros2025surrogate} at the University Jaume I in Spain; at INESCTEC ISEP~\cite{INESC-TEC_CRAS_lab} in Portugal; LABUST~\cite{ferreira2024remote} at the University of Zagreb, Croatia; and at Aalborg University~\cite{AAU_UnderwaterTechnology} and the \gls{asta} at the \gls{dtu}~\cite{asta-dtu}.

\Gls{mocap} systems are used for high-quality data collection and tracking both underwater \cite{sintef-ocean-lab, brest, umich, UMD_NBRF, DFKI_MaritimeInfrastructure} and on air \cite{sintef-ocean-lab, TCOMS_RD, brest, umich, asta-dtu}. However, of these facilities, only three include systems in both domains \cite{sintef-ocean-lab, brest, umich} as the Marinarium does, creating an ideal setup for multi-domain collaborative robotics development. 

Concerning controllable water conditions, only the largest facilities among the above~\cite{sintef-ocean-lab, TCOMS_RD, brest} can house wave and/or wind generators for realistic seafaring conditions. Current control, on the other hand, is easier to achieve and is therefore a more common feature in smaller ones ~\cite{aquatron_pool_tank, monros2025surrogate,asta-dtu, gunnarson2024fish}. Water turbidity regulation is useful to mimic realistic visibility conditions for testing of vision-based navigation systems. However, the performance of underwater \gls{mocap} systems is highly dependent on the clarity of the water volume and therefore only a smaller dedicated tank \cite{DFKI_MaritimeInfrastructure} offers this feature. 

A further characteristic among water tanks is suitability for human diving, summarized in the Divers column of \cref{tab:uwf}. Most robotics-oriented facilities at a similar scale either explicitly disallow diving or do not disclose this information, often because their water volumes are not maintained to required water standards for humans. The Marinarium however was designed to support diver-in-the-loop experimentation both in size, water quality and pool accessibility and supervision, as illustrated in \cref{fig:carl}.

Focusing on space robotics, the larger Neutral Buoyancy Research Facility (NBRF) at the University of Maryland~\cite{UMD_NBRF} was originally built to study orbital operations to support NASA, but is now also equipped for \gls{uw} robotics research. Its counterpart in Europe is the test facility at DFKI Bremen~\cite{DFKI_MaritimeInfrastructure}, intended for both \gls{uw} and space robotics ~\cite{koch2020underwater}. The neutral buoyancy labs at NASA~\cite{jairala2012eva} and \gls{esa}~\cite{marciacq2009crew} focus primarily on astronaut training. To the best of our knowledge, none of the small or medium-sized facilities incorporate infrastructure for combined underwater and space robotics research.

Regarding digital representations of the facilities, while several of them have accompanying simulation environments modeling some aspects of their water tanks~\cite{lindsay2022collaboration, ferreira2024remote, DFKI_MaritimeInfrastructure, jairala2012eva, CIRS},
the only published literature found that demonstrates a digital twin of any of the indoor \gls{uw} research facilities is CIRTESU~\cite{lopez2025unity}. 

In addition to the specialized facilities listed, examples of towing tanks~\cite{lin2026intelligent, xu2022integrated, song2024turtlmap}, and swimming pools~\cite{brantner2021controlling, xanthidis2020navigation, yang2024gliding} employed ad-hoc for \gls{uw} robotics testing are common in the literature, which is a sign of both the scarcity of these specialized facilities and their value to this research.

Among the facilities surveyed in \cref{tab:uwf}, the Marinarium is, to the best of our knowledge, the only one that combines all of the following: underwater and above-water \Gls{mocap} in a single facility, a retractable roof exposing surface and air operations to natural weather while maintaining controlled water temperature, a published digital-twin workflow, diver-suitable conditions, and a tightly coupled space robotics laboratory.  
In the following sections, we present the related work corresponding to the four sets of experiments made possible by the Marinarium.

\subsection{UNDERWATER VEHICLE DYNAMICS MODELING}
Vehicle dynamics models are an essential building block for control design, state estimation or simulation techniques. The modeling of maritime vehicles has traditionally relied on \gls{cfd} methods or data from resistance and self-propulsion tests from scaled-down models in a towing tank~\cite{lopes2025resistance}.  
Data-driven techniques are becoming increasingly popular as a lighter and more cost-effective alternative to these methods~\cite{amer2025modellingunderwatervehiclesusing}. 
Ground truth data is difficult to obtain in open water experimentation, but is readily available in controlled tank facilities with underwater \Gls{mocap}. 

Regarding first-principle methods, the study in~\cite{Wu2018BlueROV2} estimated hydrodynamic coefficients from basic geometry and mass properties for a BlueROV2, whereas the work in~\cite{vonBenzon2022BlueROV2} carried out an extensive system-identification campaign to obtain more reliable parameters. Both works highlight that accurate added-mass and damping estimation requires controlled experiments and is difficult to maintain across vehicle configurations, motivating data-driven alternatives.

The review in~\cite{christensen2022recent} surveys \gls{ai} methods for underwater navigation and control and underlines model learning as a key trend. Early applied work such as~\cite{wehbe2017learning} used Support Vector Regression to capture coupled \gls{auv} dynamics from onboard logs. More recently, the approach in~\cite{amer2025modellingunderwatervehiclesusing} proposed a \gls{pinc} for BlueROV2 modeling and, in~\cite{dynamic_GP_brov2}, a dynamic-forgetting \gls{gp} residual model embedded in \gls{mpc}, which augments, rather than replacing, a physics baseline. As an alternative, Koopman operator approaches, like \gls{edmdc}, provide linear, control-coherent surrogates; e.g.,~\cite{koopman_manip} demonstrated accurate predictive control for a nonlinear robotic manipulator. 
The controlled instrumented environment provided by the Marinarium enables the collection of the dense, low-noise state-input datasets required for such data-driven operator-theoretic system identification. To our knowledge, Koopman-based \gls{edmdc} identification has not previously been reported for full 6-DoF underwater vehicle dynamics at this prediction horizon.

\subsection{MULTI-DOMAIN HETEROGENEOUS MARITIME ROBOTICS}
Multi-domain robotic systems involve coordination between autonomous platforms in more than one of the following domains: underwater, water's surface, land, air, and space~\cite{mcconnell2025above, dos2022cross}. 
This type of system is increasingly important offshore because seaborne transportation accounts for more than 80\% of global trade volume~\cite{trade}, and the digitalization and automation of maritime activities, environmental monitoring, and search and rescue operations are relying more heavily on fleets of heterogeneous autonomous platforms \cite{shkurti2012multi, nomikos2022survey, li2023survey}. Traditionally, given their operational complexity, these types of multi-robot systems have been developed and tested in simulation~\cite{anand2021openuav}, directly offshore \cite{shkurti2012multi, mcconnell2025above}, or via a combination of both \cite{nomikos2022survey, li2023survey}. 
Further testing in controlled facilities, as an intermediate step, can save costly field testing time and speed up development. 
However, there are not many examples of this type of indoor experimental setup~\cite{lindsay2022collaboration}.
We showcase how the Marinarium provides a unique multi-domain arena for this type of experimental development through the execution of a multi-domain mission with a heterogeneous fleet of robots.

\subsection{UNDERWATER SIM2REAL GAP BRIDGING}
Although state-of-the-art simulators still fail to fully capture real-world physics accurately (the so-called simulation-to-real gap), the latest advances in sim2real bridging techniques \cite{hofer2021sim2real} have demonstrated the feasibility of successfully deploying control policies developed entirely in simulation directly into real-world hardware \cite{aljalbout2025reality}.
However, the intrinsically higher complexity of modeling hydrodynamics accurately has hindered achieving a similar level of success in bridging the sim2real gap in the underwater domain. The performance deterioration caused by this gap can be seen in the fact that many underwater reinforcement learning works remain in the simulation realm \cite{chang2025learning, lidtke.2024}, or are tested in real hardware only under very limited conditions \cite{cai2025learning, sufan.2025}. 
A practical alternative to more accurate but computationally expensive hydrodynamic simulators is the use of residual dynamics models to complement the simulated robot dynamics, as demonstrated in \cite{kaufmann2023champion, gao2024sim}. Residual models capture the discrepancy between simulated and real-world vehicle motion, and subsequently enhance the fidelity of the simulator. However, learning these residuals requires access to high-accuracy, real-world motion data paired with the simulated vehicle to be enhanced, which has prevented this technique from being applied to underwater simulators. To bridge this gap, in this work we use data from the underwater \Gls{mocap} system in the Marinarium together with the Marinarium's digital twin in the SMaRCSim and demonstrate how residual dynamics regression can be applied to improve underwater dynamics simulation.

\subsection{SPACE VALIDATION UNDERWATER}
Neutral buoyancy environments have long served as microgravity surrogates for space research, supporting both astronaut training and robotic testing in facilities such as NASA's Neutral Buoyancy Laboratory~\cite{jairala2012eva} and the University of Maryland's Neutral Buoyancy Research Facility~\cite{akin_robotic_capabilities}. Several underwater spacecraft analogs were developed in these settings, including SCAMP~\cite{miller_attitude_2000}, Ranger~\cite{carignan_reaction_stabilization}, and EUCLID~\cite{nasa_vertical_2014}, demonstrating that underwater vehicles can emulate free-flying spacecraft behavior for inspection, servicing, and control studies. Despite their early success, this line of research has seen limited continuation. 
Simultaneously, planar air-bearing testbeds have become the standard for ground-based emulation of spacecraft dynamics. 
One such air-bearing testbed is the \gls{atmos}~\cite{roque2025towards}, located at \gls{kth}.

To the best of the authors' knowledge, NASA \gls{jsc} is the only other facility housing both an \gls{uw} lab and a precision air bearing floor for space experiments~\cite{NASA_facilities}. The Marinarium is smaller in size, but comes with \gls{uw} \gls{mocap} for ground truth positioning. Furthermore, the Marinarium and its robots have been designed to run the same software stack as \gls{atmos} to facilitate parallel space-simulation experiments between the two facilities.
Each emulation facility has different strengths for simulating space missions, such as the Marinarium's ability to simulate 3D motions and \gls{atmos}'s ability to simulate drag-free planar motion.  
Together, these facilities provide a unique testing environment.\\

In the remainder of this work, we introduce the Marinarium's capabilities in \cref{sec:facility}, and then showcase, through a set of experiments, four key open research areas within field robotics that the Marinarium facilitates research in, namely: i) data-driven system identification of underwater vehicles in \cref{sec:sysid}, ii) sim2real gap bridging for underwater simulation in \cref{sec:sim2real}, iii) multi-domain maritime robotics testing under controlled conditions in \cref{sec:multi_domain} and iv) validation of space robotics using underwater surrogates in \cref{sec:space}.

\begin{figure*}[t]
    \centering
    \includegraphics[width=0.99\textwidth]{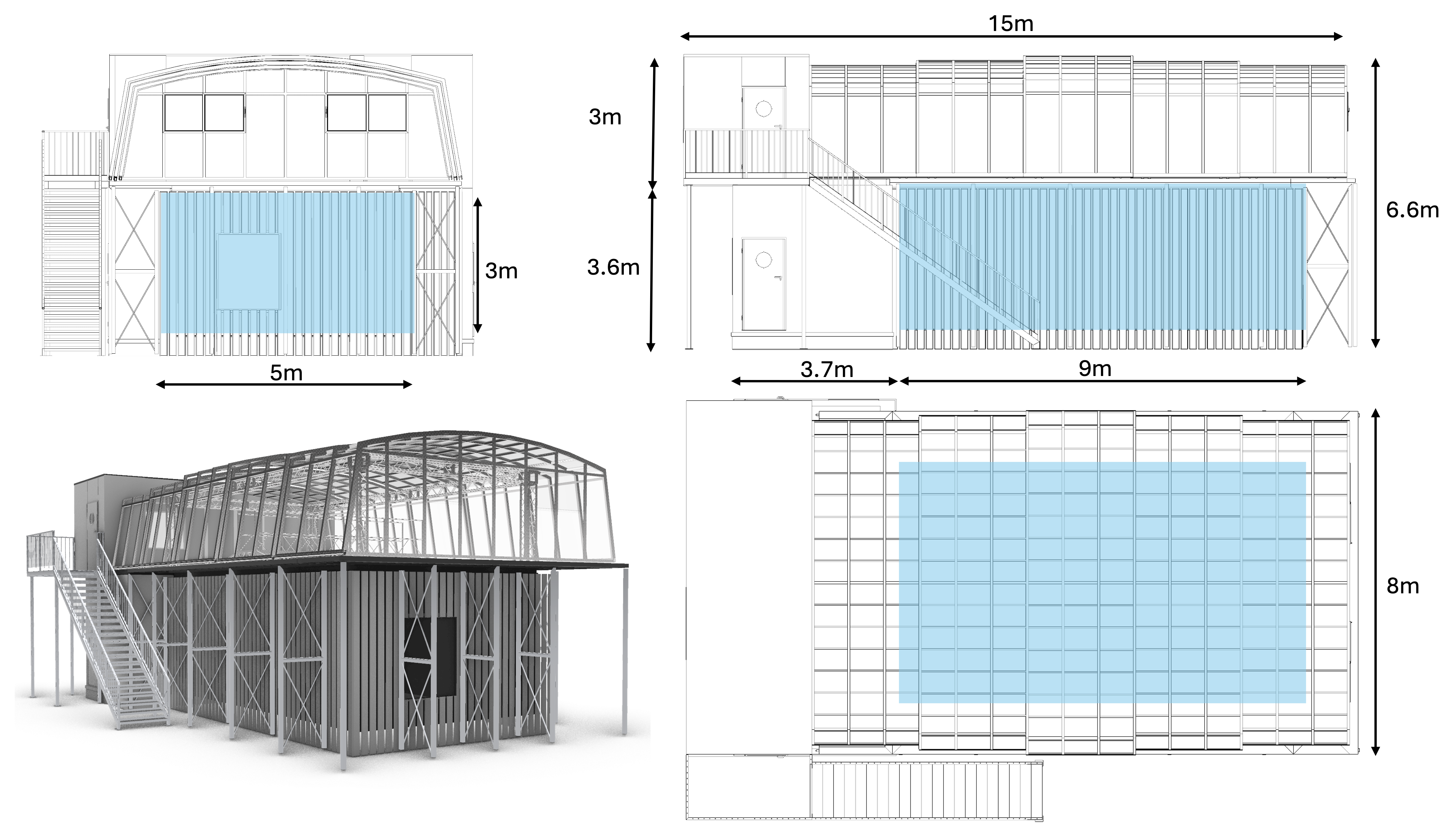}
    \caption{Plans and CAD model of the Marinarium with the dome closed. The structure is built combining pre-fabricated modules, a floating terrace at water surface level and a retractable roof, to form a stand-alone facility. The \unit[9$\times$5$\times$3]{m} water basin volume has been depicted in blue for visualization purposes.}
    \label{fig:marinarium_drawing}
\end{figure*}

\section{THE MARINARIUM FACILITY}
\label{sec:facility}
The Marinarium has been designed with modularity and resource-efficiency at its core. To this end, it is comprised of pre-built sections that can be assembled, disassembled, and modified as needed. It consists of three main modules: i) an overground \unit[3]{m} deep water basin of \unit[9$\times$5]{m}, ii) two stacked office modules with direct access to the surface of the pool from the top module and large underwater windows into the water volume in the bottom module, and iii) a freestanding terrace at water surface level with a retractable roof. These three components have been combined to create a standalone facility that can house equipment and carry out long-term experiments under repeatable and stable conditions. The CAD model and main plans can be seen in \cref{fig:marinarium_drawing}. Owing to these design choices, the overall construction costs of the facility (not including the \gls{mocap} systems) is estimated to be under 4M Swedish krona. 

\subsection{MULTI-DOMAIN OPERATING WORKSPACE}
The Marinarium water basin has a total volume of \unit[135]{m$^3$}. A Qualisys underwater \gls{mocap} system with 8 Arqus cameras and 1 Miqus video camera has been installed just below the surface to provide full coverage of the basin. The camera setup can be seen in \cref{fig:rov_mocap}. A truss system has been set up along the perimeter of the basin to enclose a \unit[9$\times$5$\times$2]{m} area over the water that can be fully confined and monitored. Alternatively, this volume can be expanded into the open air through the retractable roof, allowing experiments to be carried out under real meteorological conditions closer to the real world. For tracking in the volume over the water surface, an Optitrack \gls{mocap} system with up to 16 Flex 13 cameras can be set up and reconfigured on the truss system. The Qualisys and Optitrack MoCap systems are integrated through \gls{ros}~2.

\subsection{UNDERWATER ACOUSTIC PERFORMANCE}
\begin{figure}[h]
    \centering
    \includegraphics[width=0.99\linewidth]{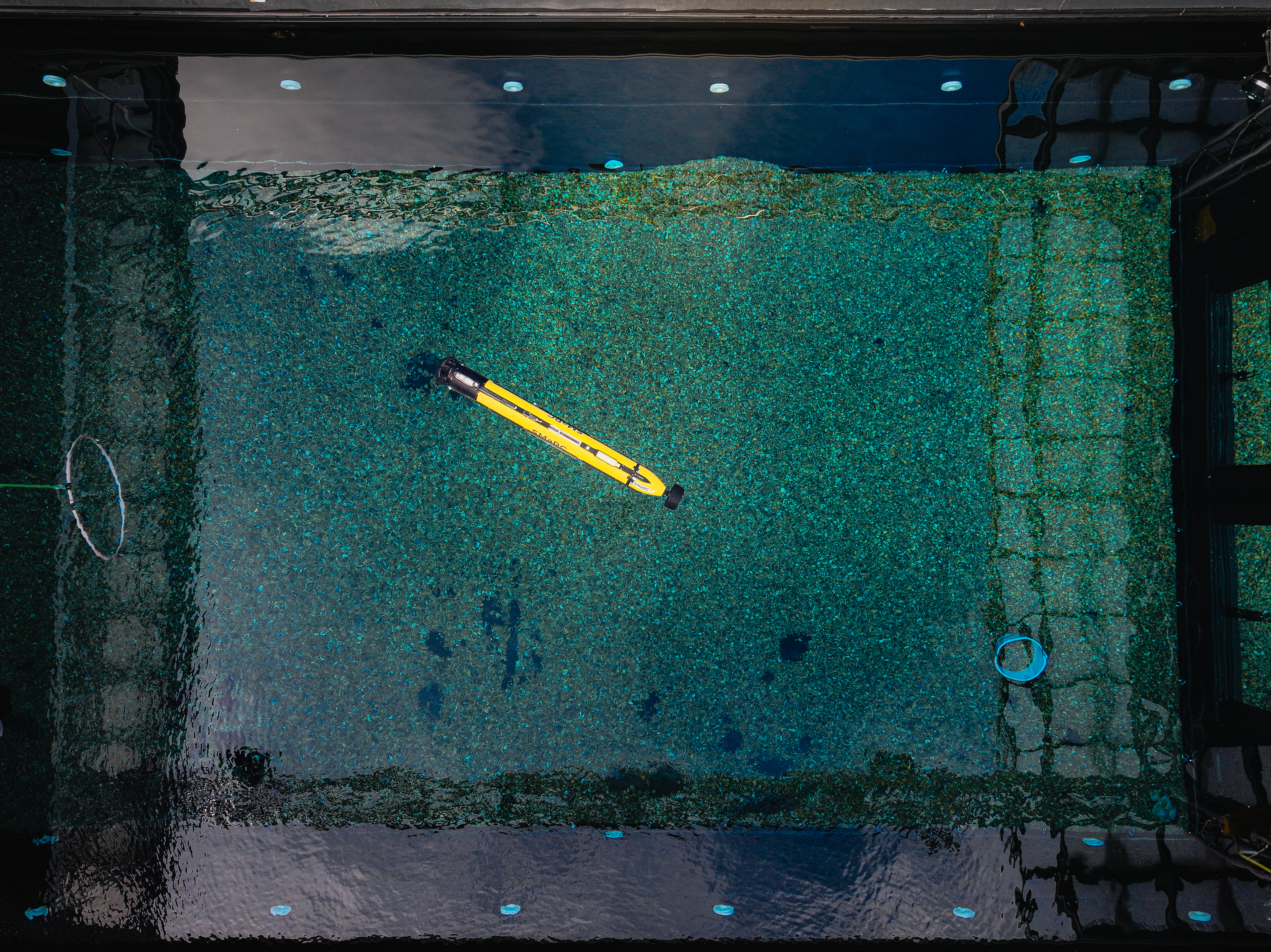}
    \caption{Partial aerial view of the Marinarium water basin. 3 tons of sea gravel of \unit[8-16]{mm} radius have been laid on the floor to reduce acoustic multipath and enable the operation of low-frequency acoustic equipment.}
    \label{fig:water_basin}
\end{figure}

As electromagnetic radiation is quickly absorbed underwater, maritime field robots often use acoustic communication, which is difficult to use in water tanks due to multipath effects. 
In the Marinarium, with a water basin consisting of a steel skeleton supporting a \unit[20]{mm} PE100 RC liner that houses \unit[135]{t} of water, we address this by covering the bottom with gravel. 
The bottom is covered with \unit[3]{t} of yellow-mottled sea gravel of \unit[8-16]{mm} radius (\cref{fig:water_basin}), which scatter acoustic reflections from sonar instruments and reduce multipathing in low-frequency instruments in the tank. 
As an example, the use of these stones has enabled a transmission rate of $70\%$ with Delphis Succorfish modems (\unit[24-32]{kHz} acoustic frequency), up from $0\%$ without stones.

\subsection{WATER TREATMENT SYSTEMS}
The tank has been instrumented to preserve water conditions within operating ranges in terms of clarity for the Qualisys \gls{mocap}, sanitation for diving work, and temperature control for infrastructure maintenance.
The equipment consists of two equivalent and independent piping subsystems for robustness. Each of them encompasses a chlorination dispenser, an \gls{uv} disinfection unit, a sand-based filter, and a heating system. 
Two heat pumps provide an actionable water temperature range of \unit[6-40]{$^{\circ}$C}. Their purpose is two-fold: i) to keep the water within the operating temperature of the Qualisys cameras, \unit[0-35]{$^{\circ}$C}, and ii) to prevent pipes from freezing during winter. Regarding sanitation, the automatic chlorination dispenser maintains the water quality within the working standards for human use, while the water clarity is maintained to optimize the \gls{mocap} measurement distances via two \unit[800]{kg} sand filters. Finally, two \gls{uv}-C cleaners are used to both disinfect the water and photodegrade the chlorine byproducts, which are known to increase water turbidity.

\subsection{DIGITAL TWIN}
The digital twin of the Marinarium facility has been implemented in the SMaRCSim simulator \cite{kartavsev2025smarcsim}, and can be seen in \cref{fig:carl}. This replica of the facility is integrated with the real infrastructure via a \gls{ros}~2-Unity bridge. Through this bridge, the real-time state of the vehicles in the environment, both from the \glspl{mocap} and their internal estimators, can be directly forwarded into the simulator. These digital representations of the real vehicles in the simulation can then be directly compared to their simulated counterparts. Equivalently, this setup enables seamless \gls{sil} testing from the simulator into the real vehicle prior integration into the real hardware.
\begin{figure}[h]
    \centering
    \includegraphics[width=0.9\linewidth]{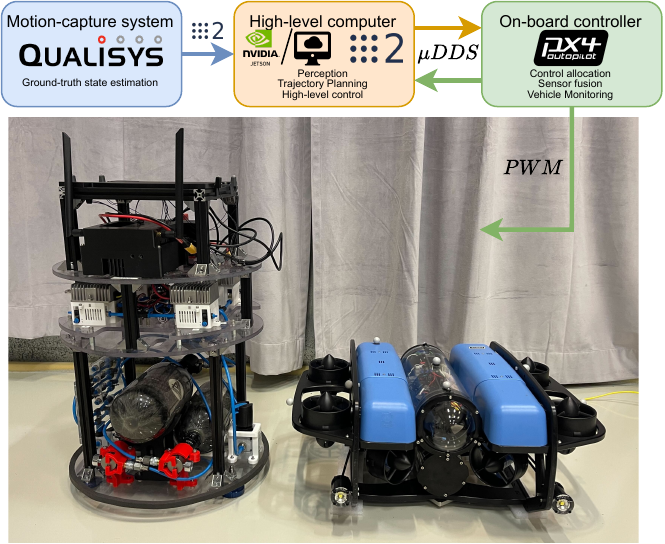}
    \caption{The \gls{atmos} free-flyer~\cite{roque2025towards}, the BlueRobotics BlueROV2 Heavy, and their identical autonomy stack. Ground truth state estimation is obtained from a Qualisys \gls{mocap} system and transferred to either an on-board Nvidia Jetson or an off-board PC via \gls{ros}~2. High-level computing and decision-making is performed in \gls{ros}~2 after which high-level control signals are sent to an onboard PixHawk microcontroller which allocates the control to the individual actuators.}
    \label{fig:atmos_brov}
\end{figure}

\subsection{INTEGRATION WITH THE KTH SPACE ROBOTICS LAB}
The Marinarium is located adjacent to the \gls{kth} space robotics laboratory \cite{roque2025towards}, within the campus of \gls{kth}. The geographic proximity allows the facilities to both share resources and be connected through an optical fiber network, enabling them to operate in the same \gls{ros}~2 domain and carry out combined experiments. 
Furthermore, in order to build an underwater surrogate platform of the \gls{atmos} free-flyer~\cite{roque2025towards} vehicles in the space robotics laboratory, a set of BlueRobotics BlueROV2 Heavy have been equipped with an identical software and control stack. Each robot is equipped with a Pixhawk 6X–Mini microcontroller running PX4~\cite{meier_px4_2015}, which performs onboard sensor fusion of inertial and external motion-capture data and computes low-level control by allocating commanded forces and torques to the actuators. PX4 provides modular configuration, robust control features, and native integration with \gls{ros}~2, ensuring consistent control logic and data handling across both platforms. All communication, supervision, and data logging are handled through \gls{ros}~2.

\section{RESEARCH AREA 1: DATA-DRIVEN \gls{sysid} FOR \gls{uw} VEHICLES}
\label{sec:sysid}
Accurate dynamics models are central to navigation, control, and simulation of underwater vehicles. They provide the predictive baseline for controller design, state estimation, and pre-deployment testing and validation. The Marinarium enables the collection of repeatable, high-fidelity experimental data for model identification and validation, allowing more accurate vehicle dynamics models to be developed and systematically evaluated. We begin by recalling the physics-based formulation commonly used in underwater robotics.

\subsection{UNDERWATER VEHICLE DYNAMICS}
The dynamics of underwater vehicles are typically described via the standard Fossen model~\cite{fossen2021handbook} as:
\begin{align}
\label{eq:fossen}
    \inertia\dot{\veltwist} + \coriolis(\veltwist)\veltwist 
    + \drag(\veltwist)\veltwist + \gravity(\pose)
    &= \wrench(\actinput), 
\end{align}
where $\pose = [x,y,z,\phi,\theta,\psi]$ is the \gls{auv} pose in the inertial frame and $\veltwist = [u, v, w, p, q, r]$ is the body-frame velocity twist.
The inertia matrix is given by $\inertia = \inertia_{RB} + \inertia_{A}$, which combines the rigid-body and added-mass contributions. Similarly, the Coriolis and centripetal terms are grouped in $\coriolis(\veltwist) = \coriolis_{RB}(\veltwist) + \coriolis_A(\veltwist)$, whereas $\drag(\veltwist)$ denotes the hydrodynamic damping forces. The term $\gravity(\pose)$ captures the gravitational and buoyancy restoring forces, and $\wrench(\actinput)$ is the actuated wrench. In practice, this wrench is not directly actuated; instead, it is obtained from the individual thruster and actuator inputs, denoted as $\actinput$. 

Despite the widespread use of the Fossen model in \cref{eq:fossen}, obtaining an accurate and reliable parameterization of the model remains difficult in practice. In particular, the hydrodynamic damping coefficients in $\drag(\cdot)$ and the added-mass terms in $\inertia_A$ are challenging to identify due to strong coupling effects, nonlinear velocity dependence, and sensitivity to operating conditions~\cite{fossen2021handbook}.
Furthermore, the resulting models are sensitive to weight and buoyancy trimming, fairing, payload reconfiguration, and seemingly minor configuration choices. 
Such factors are unavoidable during real deployments, making models based on a static parameter set, e.g.,~\cite{vonBenzon2022BlueROV2}, unreliable and difficult to validate across slightly different vehicles and domains. This motivates a data-driven surrogate of \cref{eq:fossen} that (i) identifies the effective closed-loop dynamics under the as-sailed configuration, and (ii) can be rapidly re-fit after hardware changes.


\subsection{DATA-DRIVEN DYNAMICS MODELING}
\begin{figure}[h]
    \centering
    \includegraphics[width=0.95\linewidth]{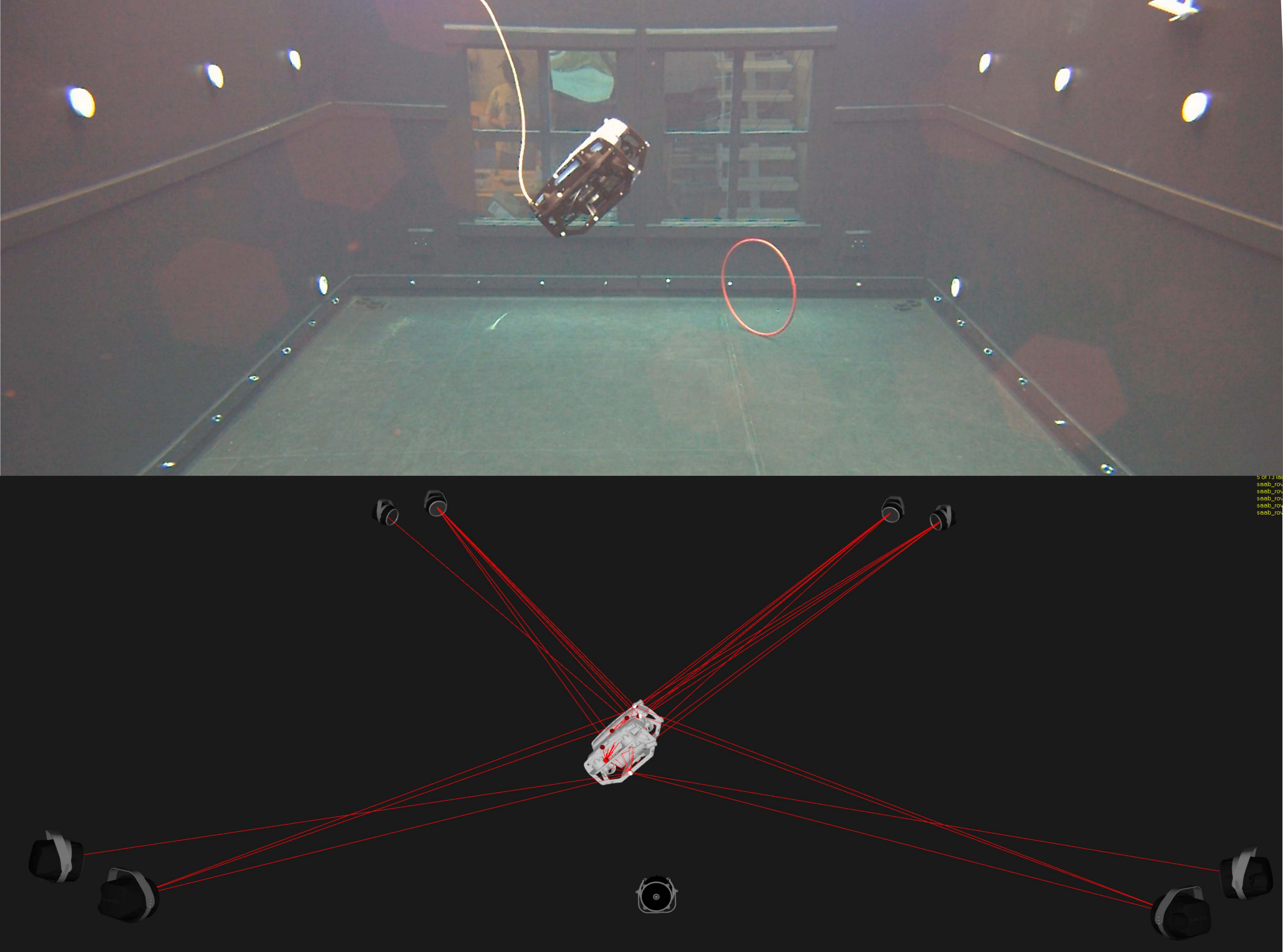}
    \caption{The BlueROV2 Heavy in the Marinarium during the data collection for sysID (top). The vehicle state $\vect{x}_k$ is provided by the Qualisys motion capture system (bottom).}
    \label{fig:rov_mocap}
\end{figure}

In this work, we propose to identify a discrete-time dynamics model of the underwater robot directly from data. The Marinarium provides a controlled and repeatable testing environment, allowing the collection of high-fidelity datasets from an instrumented BlueROV2 (\cref{fig:rov_mocap}). For this study, we recorded three datasets of more than \unit[15]{min} each, containing synchronized measurements of the full $12$-dimensional vehicle state and the $8$ thruster command channels. Some datasets contain limitations on the pitch and roll movements to ensure a fair comparison with the 4-DoF-based models, e.g.,~\cite{amer2025modellingunderwatervehiclesusing}, while others contain full 6-DoF motion.

The state vector, comprising position, orientation, and linear and angular body-frame velocities, is defined as
\[
\vect{x} \doteq [\pose^\top,\veltwist^\top]^\top \in \mathbb{R}^{12},
\]
and the input $\actinput \in [-1,1]^8$ are the individual BlueROV2 motor commands. The data for states and inputs is obtained from the underwater \gls{mocap} (pose and odometry), onboard \gls{imu} (angular rates/linear acceleration), and controller logs (thrusters). After data collection, we sort measurements by timestamp, remove duplicates and NaNs, and resample to \unit[50]{Hz} (median sample time $\samplingtime \approx \unit[0.02]{s}$), yielding more than $45823$ time-aligned samples per dataset. Unless otherwise stated, splits are chronological (first $80\%$ train, last $20\%$ test) to avoid leakage. 


\subsection{BASELINES AND MODELS}
In the following, we compare four models trained on the same dataset.
Let  $\rotmat(\phi, \theta, \psi) \in \mathbb{R}^{3\times 3}$ denote the rotation matrix from the body frame $(b$) to the inertial/navigation frame $n$, constructed from ZYX Euler angles $\eulang = [\phi, \theta, \psi]^\top$. 
We partition the state in position $\pos=[x,y,z]^\top$, euler angles $\eulang$, linear velocity $\vel=[u,v,w]^\top$, and angular velocity $\angvel =[p,q,r]^\top$. 

\paragraph*{\gls{di} in the body frame}
Under a small-angle approximation for roll and pitch, the Euler-angle kinematics simplify to
\begin{equation}
    \dot{\eulang} \approx \angvel
\end{equation}
which yields a first-order attitude update and removes the trigonometric coupling between angular rates and Euler-angle derivatives.
In this regime, the vehicle’s translational and rotational dynamics can be approximated as body-frame double integrators driven by the thruster inputs.
We model the input–acceleration relationship by estimating constant gain matrices
\[K_{\mathrm{lin}},K_{\mathrm{ang}} \in \mathbb{R}^{3\times 8}\]
using ridge-regularized least squares (Tikhonov regularization~\cite{hoerl1970ridge}). These matrices map the 8 thruster commands to linear and angular accelerations:
\[
\dot{\vel} \approx K_{\mathrm{lin}} \actinput, \quad
\dot{\angvel} \approx K_{\mathrm{ang}} \actinput,
\]
where the Coriolis term is missing since only small angular velocities are considered. Using an explicit forward Euler integration step  with sampling time $\samplingtime$, the resulting discrete-time dynamics become
\begin{align}
\pos_{k+1} &= \pos_k + \samplingtime\, \rotmat(\phi_k,\theta_k,\psi_k)\,\vel_k, \\
\eulang_{k+1} &= \eulang_k + \samplingtime \, \angvel_k.
\\
\vel_{k+1} &= \vel_k + \samplingtime\, (K_{\mathrm{lin}} \actinput_k), \\
\angvel_{k+1} &= \angvel_k + \samplingtime\, (K_{\mathrm{ang}} \actinput_k), 
\end{align}

\paragraph*{Fossen (BlueROV2)} 
We adopt the BlueROV2 dynamics model from~\cite{vonBenzon2022BlueROV2}, using the nominal parameters reported therein. The continuous-time equations are integrated using a single explicit Euler step with sampling time~$\samplingtime$. No additional tuning or parameter adaptation is performed

\paragraph*{\glsentryfull{pinc}}
Using the approach described in~\cite{amer2025modellingunderwatervehiclesusing}, we trained a network with our dataset using $1000$ epochs and a loss of $1.9208$. The loss plateaued early despite using the published hyperparameters; we retain this baseline to document that off-the-shelf learning-based models do not automatically transfer to our tank dataset without further tuning.

\paragraph*{Koopman \gls{edmdc} with \gls{rbf} dictionary (ours)}
Following~\cite{koopman_brunton}, we lift the state to a higher-dimensional \emph{observable space} via a dictionary of functions $\theta(\cdot)$:
\begin{equation}
    \vect{z} \doteq \theta(\vect{x}) = \begin{bmatrix} \vect{x} \\ \varphi_1(\vect{x}) \\ \vdots \\ \varphi_K(\vect{x})\end{bmatrix}
\in \mathbb{R}^{n+K},
\end{equation} 
Here, the \emph{observable space} refers to the space of functions of the state that we choose as basis. By lifting the state into this space, nonlinear dynamics in the original state can be approximated as linear dynamics in $\vect{z}$.

The additional coordinates $\varphi_i(\vect{x})$ are Gaussian \glspl{rbf}:
\begin{equation}\label{eq:gaussian_rbf}
\varphi_i(\vect{x}) = \exp \big(-\gamma \|\vect{x}-\vect{c}_i\|_2^2\big),
\end{equation}
where $K$ is the dictionary size, $\gamma>0$ is a width parameter, and the centers $\{\vect{c}_i\}_{i=1}^K$ are selected from the training data using $k$-means clustering (details later).
The dynamics in the lifted space are modeled as a linear, control-affine system
\begin{equation}
\vect{z}_{k+1} = A \vect{z}_k + B \vect{u}_k, \qquad \vect{x}_k = C \vect{z}_k,
\label{eq:koopman_linear}
\end{equation}
with $C$ typically chosen to extract the original state coordinates from the lifted vector, and is typically referred to as \textit{decoder} matrix. The matrices $A$ and $B$ are estimated via ridge-regularized least squares:
\begin{equation} \label{eq:koopman_min}
\min_{A,B} \big\|\Theta(Y) - [A\;\;B]\,[\Theta(X)^\top\;\;U^\top]^\top \big\|_F^2,
\end{equation}
where $\| \cdot \|_F$ denotes the Frobenius norm, matrices $\Theta(X) = [\theta(x_0), \theta(x_1), \dots, \theta(x_{T-1})]$, $U = [u_0, u_1, \dots, u_{T-1}]$ and $\Theta(Y) = [\theta(x_1), \theta(x_2), \dots, \theta(x_{T})]$. \cref{eq:koopman_min} also admits the closed-form solution
\begin{equation}
[A\;\;B] = \Theta(Y) G^\top \big(GG^\top + \lambda I\big)^{-1},
\label{eq:ridge_solve}
\end{equation}
with $G = [\Theta(X)^\top\;U^\top]^\top$ and ridge parameter $\lambda>0$. A simple decoder matrix $C=[I_{12}\;0]$ is used, i.e., the first $12$ lifted coordinates reconstruct $\vect{x}$, since using a learned decoder produced similar results in our tests. The full pipeline is shown in \cref{alg:edmdc_rbf}.

\paragraph*{The advantages of \glspl{rbf}}
Gaussian \glspl{rbf} yield a universal and smooth basis on compact sets since linear combinations of \glspl{rbf} are dense in the space of continuous functions defined on $\mathcal{X}$, denoted by $\mathcal{C}(\mathcal{X}) \doteq
\{f : \mathcal{X} \rightarrow \mathbb{R} \mid f \text{ is continuous} \}$, and can therefore approximate polynomials and other smooth nonlinearities to arbitrary accuracy~\cite{wendland2005scattered}. Practically, they provide locality, i.e., each $\varphi_i$ characterizes a nonlinearity in the state dynamics in a local neighborhood of the corresponding center $\vect{c}_i$. Such a feature helps to closely capture configuration-dependent effects like tether drag without hand-crafting cross terms. \glspl{rbf} features are also linear in parameters, enabling the closed-form ridge solve in \cref{eq:ridge_solve}.

\paragraph*{$k$-means for \gls{rbf} centers}
Given training states $\{\vect{x}^{(j)}\}_{j=1}^N$, the $k$-means centers $\{\vect{c}_i\}_{i=1}^K$ minimize within-cluster squared error:
\begin{equation}
\min_{\{\vect{c}_i\},\,\{\mathcal{C}_i\}} 
\sum_{i=1}^{K}\;\sum_{\vect{x}^{(j)}\in \mathcal{C}_i} 
\left\|\vect{x}^{(j)} - \vect{c}_i\right\|_2^2,
\quad
\vect{c}_i = \frac{1}{|\mathcal{C}_i|}\sum_{\vect{x}^{(j)}\in \mathcal{C}_i} \vect{x}^{(j)},
\label{eq:kmeans}
\end{equation}
where $\mathcal{C}_i$ is the set of points assigned to center $\vect{c}_i$ and the random seed is fixed for reproducibility.

\begin{algorithm}[t]
\caption{\gls{edmdc}--\gls{rbf} identification for BlueROV2}
\label{alg:edmdc_rbf}
\begin{algorithmic}[1]
\STATE \textbf{Input:} Snapshot tuples $\{(\vect{x}_k,\vect{u}_k,\vect{x}_{k+1})\}_{k=0}^{T-1}$, dictionary size $K$, \gls{rbf} width $\gamma$, ridge $\lambda$.
\STATE \textbf{Centers (k-means):} Run $k$-means \cref{eq:kmeans} with $K$ clusters on $\{\vect{x}_k\}$ to obtain centers $C=\{\vect{c}_i\}_{i=1}^K$.
\STATE \textbf{Lift:} Define $\theta(\vect{x}) = [\vect{x}^\top,\;\varphi_1(\vect{x}),\dots,\varphi_K(\vect{x})]^\top$ using $\varphi_i(\vect{x})$ from \cref{eq:gaussian_rbf}.
\STATE \textbf{Assemble:} $\Theta(X) = [\theta(\vect{x}_0),\dots,\theta(\vect{x}_{T-1})]$, $\Theta(Y)=[\theta(\vect{x}_1),\dots,\theta(\vect{x}_T)]$, and $U=[\vect{u}_0,\dots,\vect{u}_{T-1}]$.
\STATE \textbf{Ridge solve:} Form $G$ and compute $A, B$ with \cref{eq:ridge_solve}.
\STATE \textbf{Decoder:} Set $C=[I_{12}\;0]$ (or fit a linear decoder).
\STATE \textbf{Return:} $(A,B,C)$ from \cref{eq:koopman_linear}, centers $C$.
\end{algorithmic}
\end{algorithm}

\subsection{EVALUATION PROTOCOL} \label{sec:sysid-eval}
We report endpoint $H$-step \gls{rmse} computed by rolling each model open loop (i.e., without re-initialization) under the recorded inputs:
\begin{equation} \label{eq:RMSE_H}
\mathrm{RMSE}_H \doteq
\sqrt{\frac{1}{(N{-}H)\,n}\sum_{k=0}^{N-H-1}\!\!\big\|\vect{x}_{k+H} - \hat{\vect{x}}_{k+H|k}\big\|_2^2},
\end{equation}
for $H \in \{1,10,100\}$, state dimension $n{=}12$, $N=9165$ being the total number of data points, and Euclidean norm $\|\cdot\|_2$. Note that \cref{eq:RMSE_H} calculates the average square error for all possible forward trajectories of size $H$ started at the beginning of the test set until its near end, $N-H-1$. For our test size of more than \unit[3]{min} at \unit[50]{Hz}, this means we are averaging over more than $9000$ diverse trajectories incorporating the widest possible variety of movements and starting conditions to reduce the model bias.

To ensure a fair comparison, all open-loop rollouts are performed with one-step Euler integration. An RK4-version was also analyzed with no noticeable differences. Hyperparameters for the Koopman model, $(K,\gamma,\lambda)$, were tuned via a small grid search on a subset of the training data and then fixed for the test set. The final selected values are $K=500$, $\gamma=3.0$, and $\lambda=0.1$.

\begin{figure}[h]
    \centering
    \includegraphics[width=\linewidth]{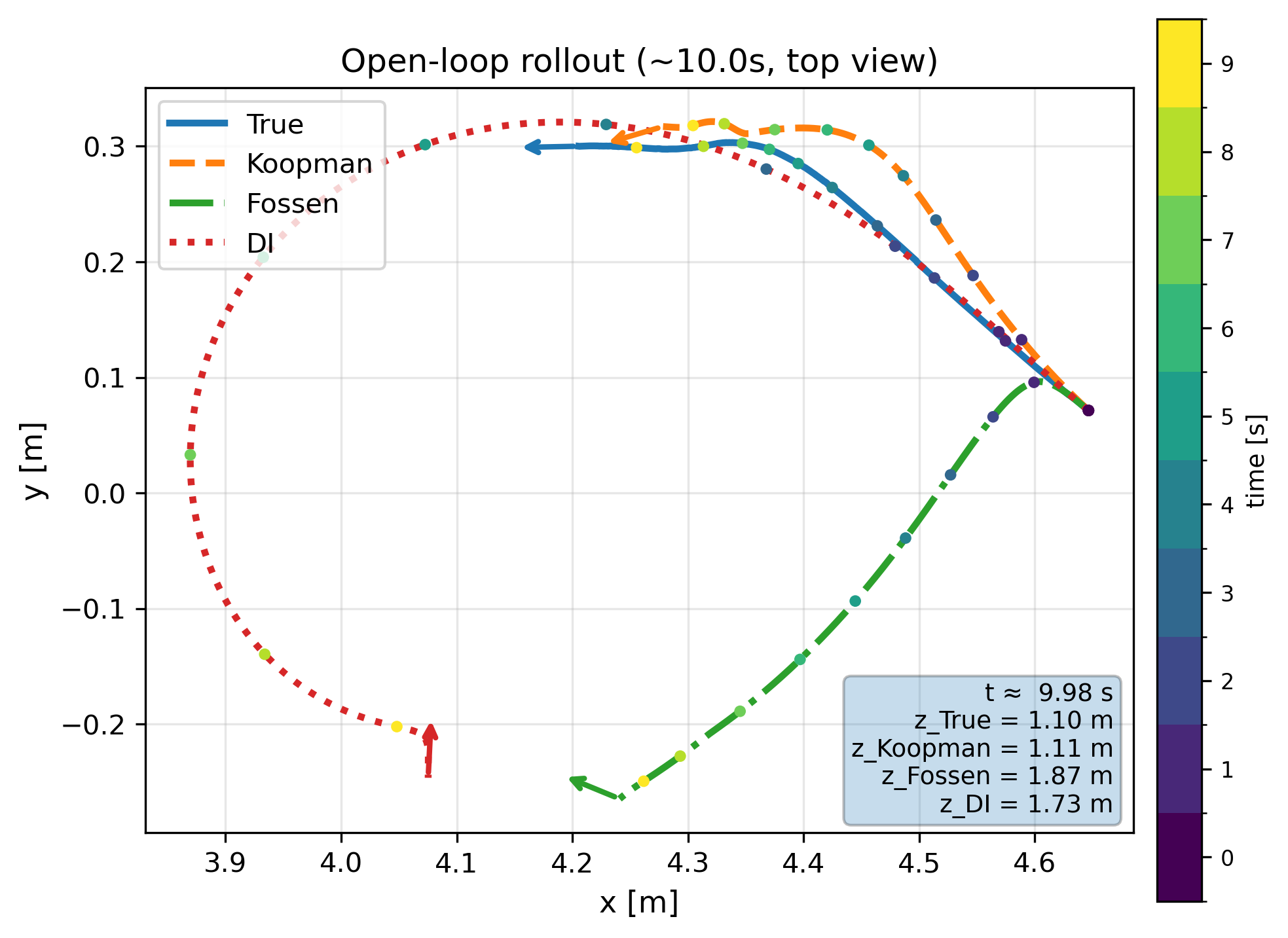}
    \caption{Recorded open-loop tank run (top view \& depth) for \unit[10]{s}: ground truth vs. Koopman, physics and \gls{di} rollouts under identical inputs.}
    \label{fig:fourup_xy}
\end{figure}
\subsection{RESULTS AND DISCUSSION}
\Cref{tab:uw_koopman_rmse} summarizes the test-set performance. The Koopman model achieves the best accuracy at all prediction horizons. The \gls{di} baseline remains competitive at $H{=}1$ and $H{=}10$, but exhibits more pronounced drift at $H{=}100$. The physics-based Fossen model from~\cite{vonBenzon2022BlueROV2} starts in a comparable range, but its accuracy degrades more rapidly, consistently with a small parameter mismatch under the specific vehicle configuration used in the tank. Finally, with the current hyperparameter choice, \gls{pinc} does not appear to learn a usable model of the underwater vehicle dynamics.

\begin{table}[t]
\centering
\caption{Endpoint \gls{rmse} \cref{eq:RMSE_H} on recorded tank data for the full test data size of more than 9000 trajectories (lower is better).}
\label{tab:uw_koopman_rmse}
\begin{tabular}{lccc}
\hline
\textbf{Model} & \textbf{1-step} & \textbf{10-step} & \textbf{100-step} \\
\hline
Koopman (\gls{edmdc}--\gls{rbf})      & 0.0629 & 0.0831 & 0.1859 \\
\glsentryfull{di}    & 0.0784 & 0.1088 & 0.4683 \\
Fossen (BlueROV2)         & 0.0765 & 0.2122 & 0.5788 \\
\gls{pinc} (ResDNN)             & 8.7886 & 9.1550 & 9.1639 \\
\hline
\end{tabular}
\end{table}

\Cref{fig:fourup_xy} reports a top-view and depth ($z$) comparison over \unit[10]{s} of open-loop rollout. For readability, the remaining six linear and angular velocity components, as well as roll and pitch, are not shown, although they are all included in the \gls{rmse} values reported in \cref{tab:uw_koopman_rmse}. It is also worth noting that the comparison is particularly demanding, as it considers significantly longer open-loop rollouts than previous works, which typically reported less than \unit[2]{s} of \gls{ivp} evolution~\cite{amer2025modellingunderwatervehiclesusing}, that is, before the real and predicted trajectories diverge substantially $(\geq \unit[0.1]{m})$. Qualitatively, the Koopman model remains closest to the ground truth while also preserving the heading evolution. The \gls{di} model follows reasonably well, but overestimates the forward speed and curvature during long yaw transients and provides a poorer estimate of the depth. The physics-based rollout initially follows the measured trajectory, but progressively diverges under the same input sequence. The \gls{pinc} results are omitted from the figure due to their poor overall performance.

Although the lifted Koopman state has dimension $d=n+K>500$, both training and rollout remain computationally light: training requires only the ridge-regression solve in \cref{eq:ridge_solve}, while rollout consists of one matrix--vector multiplication and one addition per step, $\vect{z}_{k+1}=A\vect{z}_k+B\vect{u}_k$, followed by a simple decoding step, $\vect{x}_k=C\vect{z}_k$. By contrast, the physics-based baseline requires evaluating nonlinear hydrodynamic terms, thruster maps, and, when included, tether and delay effects, which increases the per-step computational cost. The measured rollout and training times reported in \cref{tab:timings_rollout}, obtained on a desktop workstation (AMD Ryzen 7 5800 CPU, 32\,GB RAM), confirm that the Koopman model remains efficient both to train and to simulate.

\begin{table}[t]
\centering
\caption{Average training and rollout time for one step prediction.}
\label{tab:timings_rollout}
\begin{tabular}{lcc}
\hline
\textbf{Model} & \textbf{Rollout time [\unit[]{s}]} & \textbf{Training time [\unit[]{s}]} \\
\hline
Koopman (\gls{edmdc}--\gls{rbf}) & 0.0800 & 4.617 \\
Fossen (BlueROV2)    & 0.1535 & - \\
\glsentryfull{di}    & 0.0105 & 0.004 \\
\gls{pinc} (ResDNN)        & 0.3265 & 482.269 \\
\hline
\end{tabular}
\end{table}

The three datasets, the identified models, and the training and comparison scripts, including the \gls{rk4} implementation, have all been open-sourced\footnote{Code available at: \url{https://github.com/ViktorNfa/bluerov2_dynamics}.}. Overall, these results highlight the value of the Marinarium as a controlled and instrumented intermediate test environment for data-driven underwater system identification, enabling repeatable model development and validation before moving to more costly and less observable field trials.


\section{RESEARCH AREA 2: MULTI-DOMAIN MARITIME ROBOTICS}
\label{sec:multi_domain}
\begin{figure}[ht]
    \centering
    \includegraphics[width=0.95\linewidth]{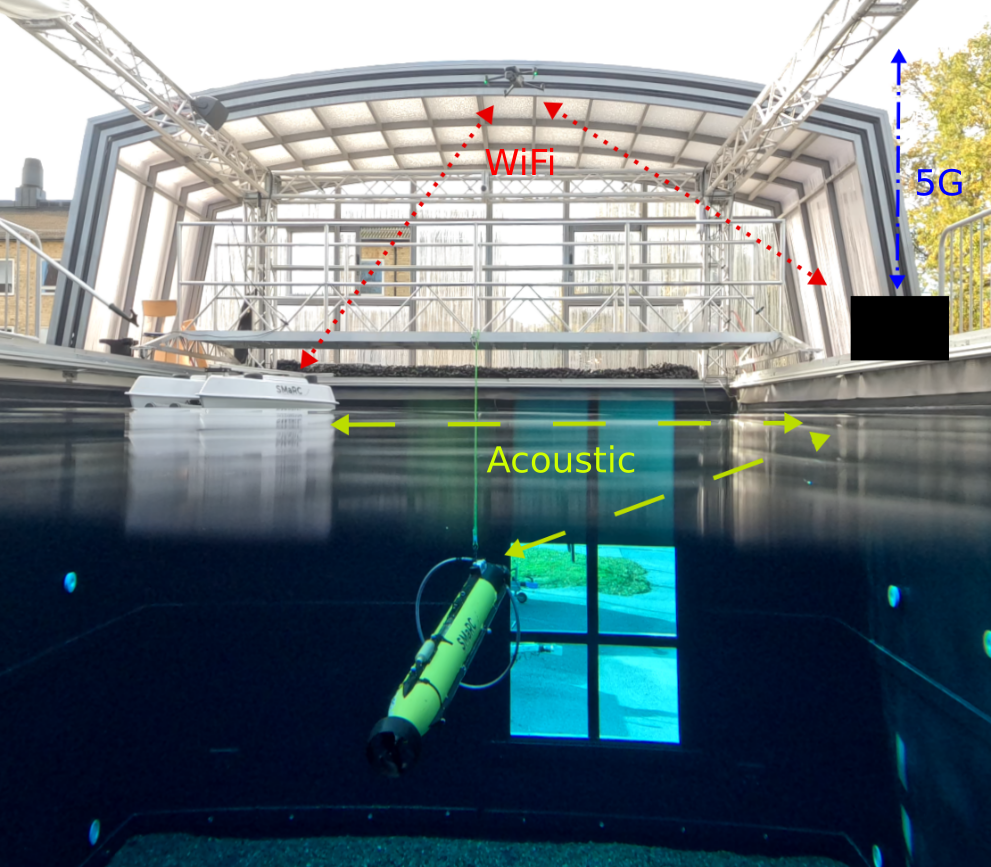}
    \caption{Example of a multidomain experimental setup with an AUV, a USV and a UAV linked through three communications channels across the aerial, surface and underwater environments. The black box represents a ground station.}
    \label{fig:multidomain}
\end{figure}

To showcase the usability of the Marinarium facility as a multi-domain maritime robotics research arena, we design and execute a rendezvous mission with a heterogeneous fleet of three robots across three environments. 
The three platforms, an \gls{auv}, an \gls{usv}, and an \gls{uav}, are depicted in \cref{fig:multidomain}, together with the ground station used for remote communication with an off-site operator. The setup is very similar to that in \cite{lindsay2022collaboration}, with the difference that untethered communication with the \gls{auv} can be achieved due to the gravel on the tank floor.


\subsection{THE HETEROGENEOUS MULTI-ROBOT FLEET}
The three vehicles and the ground station are introduced below.

\paragraph*{The SAM \gls{auv}}\label{ssec:sam} 
SAM is a small, non-holonomic, slender-type \gls{auv} developed by \gls{smarc} \cite{bhat2020cyber}. It is equipped with two counter-rotating propellers within a thrust-vectoring system for steering, and two internal actuators: i) a \gls{vbs} and ii) a \gls{lcg} system (by a moving battery pack). Its sensor suite consists of a Waterlinked A50 \gls{dvl}, a pressure sensor, a GPS module, a STIM300 \gls{imu} and SBG \gls{ahrs}, a Deep Vision side-scan sonar, and three forward-facing cameras. In the communications bay, it carries a 5G and WiFi modem as well as a Succorfish Delphis underwater acoustic modem. 
SAM's navigation architecture consists of an \gls{mpc} controller based on its dynamics model, which follows \cref{eq:fossen}, equivalent to that in \cite{bhat2023nonlinear}, together with a factor graph-based state estimator as in \cite{dorner2024smooth} to achieve autonomous waypoint navigation.

\paragraph*{The FloatSAM USV}\label{ssec:floatsam} 
FloatSAM is a small, catamaran-shaped \gls{usv} of approximately \unit[0.6$\times$0.4]{m} surface developed by \gls{smarc}. It is equipped with differential thrusters and a PixHawk 6X-Mini controller running PX4~\cite{meier_px4_2015} for low-level actuation control. Its sensor suite encompasses a Nortek \gls{dvl}, a STIM \gls{imu}, and a RTK GPS. Regarding communications, FloatSAM carries a 5G modem, a Succorfish Delphis underwater acoustic modem and a WiFi antenna.

\paragraph*{The Crazyflie UAV}\label{ssec:crazyflie} 
The Crazyflie 2.1+ is a lightweight, low-cost \gls{uav} by Bitcraze. Its native software stack implements waypoint (WP) navigation through a combination of PID-controllers and an \gls{ekf} for localization used to fuse the detections from the surface \gls{mocap}. 

\paragraph*{The Ground Station}\label{ssec:command_station} 
The ground station is a portable, waterproof communication module developed to provide a link between a remote operator and vehicles in the field. It is equipped with a Raspberry Pi 4 connected to 5G and Succorfish modems and omnidirectional WiFi. 

\subsection{ACROSS-DOMAIN COMMUNICATION LINKS}
Every platform in our setup implements \gls{ros}~2 as middleware. However, for inter-platform communication, three different links have been established consisting of a physical channel and messaging protocol, depending on the domain. These are depicted in \cref{fig:multidomain} for visualization:

\begin{itemize}
    \item WiFi/Bluetooth-ROS2 (red): all platforms within WiFi reach (or Bluetooth in the case of the Crazyradio) can communicate with each other directly through \gls{ros}~2. This includes SAM while on the surface.
    \item 5G-MQTT (blue): a NodeRED-based GUI hosted on a cloud server allows off-site users to interact with the vehicles through an MQTT link between the users and the ground station. An MQTT-ROS2 bridge in the ground station allows the user commands to be parsed into \gls{ros}~2 messages.
    \item Acoustic-NM (yellow): the \gls{nm} in \cite{fenucci2022ad} has been implemented and instantiated per vehicle to parse \gls{ros}~2 messages into low-level data frames. These are then scheduled and transferred through the acoustic link provided by the Succorfish modems between underwater platforms. 
\end{itemize}

\subsection{RESULTS AND DISCUSSION}\label{ssec:mission_pipeline} 
When commanded by an off-site operator, the \gls{auv} and the \gls{uav} receive the position of the loitering \gls{usv} tracked by the surface and underwater \glspl{mocap}, respectively. They both then proceed to approach the \gls{usv} and the mission finishes when the \gls{uav} has landed on the \gls{usv} and SAM has maneuvered underneath it. 
\begin{figure}[h]
    \centering
    \includegraphics[width=0.95\linewidth]{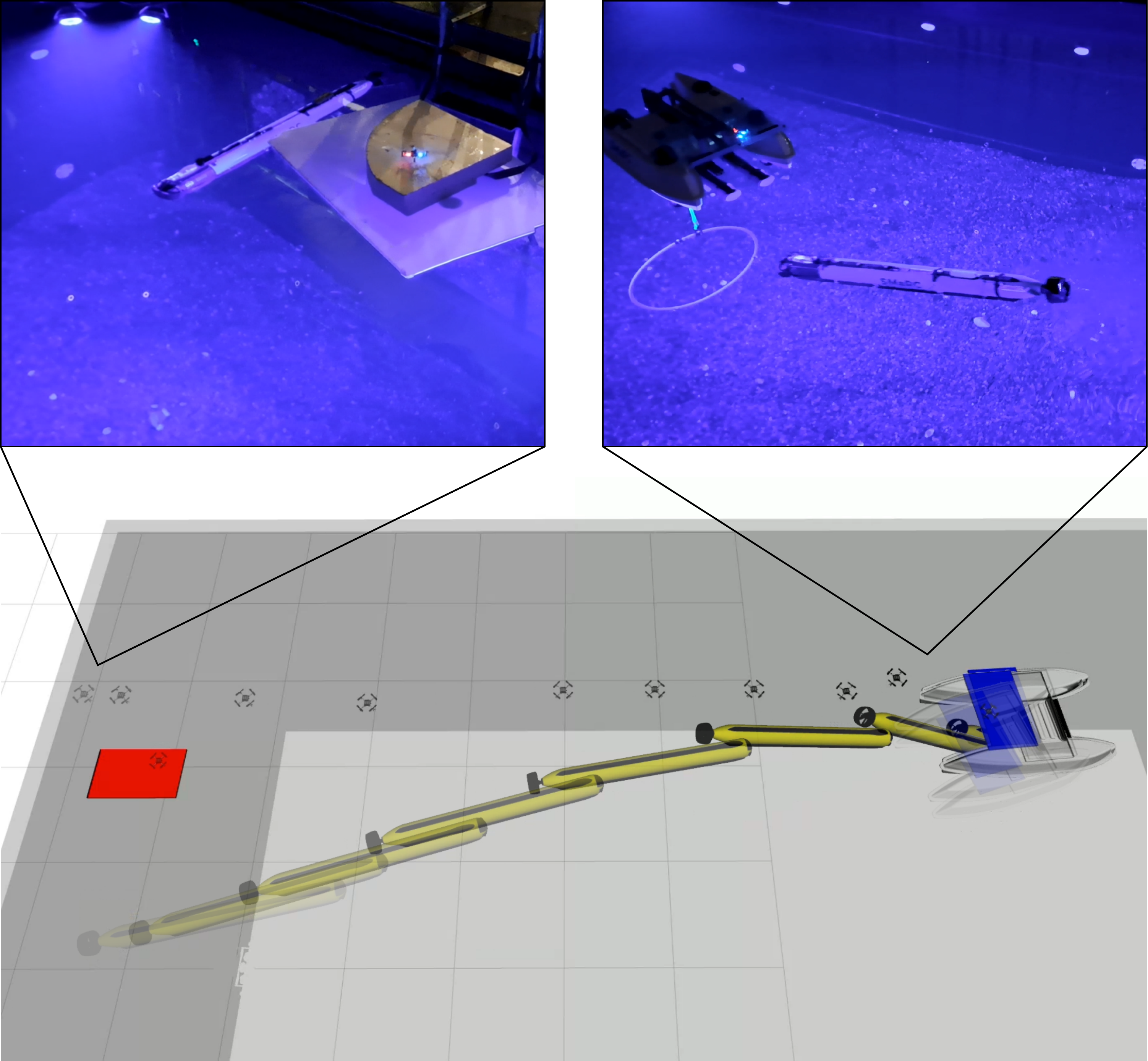}
    \caption{Time-lapse RViz visualization of the multi-domain experiment. The Crazyflie \gls{uav} takes off from the launchpad (red) and lands on the FloatSAM \gls{usv} pad (blue). In parallel, the SAM \gls{auv} dives towards FloatSAM and soft-docks underneath. On top, two real snapshots of the experiment.}
    \label{fig:rviz_multidomain}
\end{figure}

\begin{figure*}[t]
    \centering
    \includegraphics[width=0.99\textwidth]{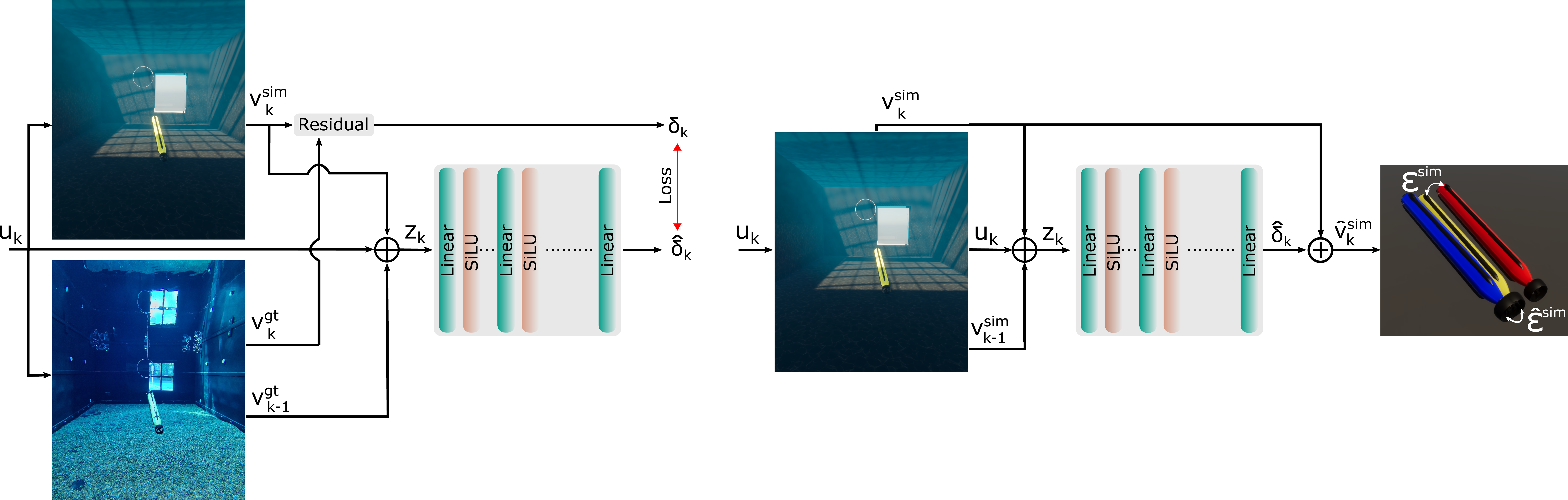}
    \caption{Left: the proposed dynamics residual learning pipeline during training with the \gls{auv} SAM. The GT and sim velocities from the \gls{mocap} and the simulator are used to compute the original residual $\tilde{\delta}_k$ over a set of \gls{auv} trajectories. A neural network is used to predict the residual $\hat{\delta}_k$ given the state and control input, concatenated in $z_k$, of the real \gls{auv}, and the difference is used to train the NN. Right: during inference, the NN is fed $z_k$ from the original simulated vehicle (yellow), with a motion error $\varepsilon^{sim}$ when compared to the real vehicle motion (red). The predicted residual is then injected, resulting in a residual-corrected model (blue) with a smaller motion error $\hat{\varepsilon}^{sim}$. }
    \label{fig:sim2real_resnet}
\end{figure*}
A timelapse of the RViz visualization of the mission is shown in \cref{fig:rviz_multidomain}. This visualization was produced by merging and replaying the rosbags collected by the vehicles and the \glspl{mocap}.
Upon reception of the start command, the \gls{uav} takes off vertically from the launch pad (red), travels at a constant altitude over the water to the coordinates of the \gls{usv}, and lands vertically on the \unit[0.5$\times$0.4]{m} pad (blue) mounted on it.
In parallel, when the \gls{auv} receives the command over the acoustic link, it navigates towards the \gls{usv} and dives beneath it until a predefined distance threshold is reached. It then empties its \gls{vbs} and surfaces, ending the mission.

The scenario presented illustrates how a full inter-domain mission with a heterogeneous fleet of vehicles can be tested in a controlled facility before field deployment, reducing operational costs, development time, and helping bridge the gap between development in simulation and field testing.

\section{RESEARCH AREA 3: UNDERWATER SIM-TO-REAL GAP BRIDGING}
\label{sec:sim2real}
Improving a robot simulator by injecting a learned residual physics model of the real robot, as in \cite{kaufmann2023champion, gao2024sim}, requires synchronized state information from both the physical platform and its simulated counterpart. The Marinarium's digital-twin infrastructure directly supports this workflow by enabling real-time comparison between experiments conducted in the facility and their simulated replicas. We therefore utilize the Marinarium digital twin in the SMaRCSim simulator \cite{kartavsev2025smarcsim}, in \cref{fig:carl}, with a \gls{ros}~2-Unity bridge.

\subsection{RESIDUAL DYNAMICS PROBLEM DEFINITION}
To demonstrate the applicability of the Marinarium to realizing this sim2real technique underwater, we aim to model the discrepancy between the observed 3D motion of the \gls{auv} SAM navigating in the Marinarium and the motion predicted by its simulation model directly from experimental data. The regressed model is then injected into the original simulation to align its behavior more closely with that of the real platform, thereby reducing the sim2real gap.
More formally, if we represent the dynamics of the SAM \gls{auv} in the SMaRCSim as $\dot{\veltwist}^{sim}_{k} = \fsam^{sim}(\pose_k, \veltwist_k, \actinput_k)$ and equivalently the true dynamics of SAM in the tank via $\fsam^{real}$, the goal is to match the dynamics of $\fsam^{real}$ by adding a residual velocity correction to $\fsam^{sim}$, named $\residual_k$.

We express the discrete-time dynamics of SAM by applying a one-step Euler discretization to $\fsam^{sim}$, which serves as the nominal physics model, i.e.
\begin{equation}
    \veltwist^{sim}_{k+1} = \veltwist^{sim}_{k} + \samplingtime \fsam^{sim}(\pose_k, \veltwist_k, \actinput_k) 
    \label{eq:nom_model}
\end{equation}

The \gls{mocap} system provides high-quality measurements of the vehicle’s velocity twist, which we use as ground truth and denote by $\gtveltwist_k$. We also denote by $\simveltwist_k$ the twist predicted by the simulation model in \cref{eq:nom_model} in the absence of residuals.
This allows us to compute the residual directly from data as $ \residual_k = \gtveltwist_{k} - \simveltwist_{k}$.

\subsection{DATASET CURATION}
To learn the residual $\residual_k$, we require a dataset of the form $\dataset = \{\gtveltwist_k, \simveltwist_k, \phi_k, \theta_k, \actinput_{0:K-1} \}$ for $k = [0... K]$. The predicted velocities $\simveltwist$ are obtained directly from the simulator by single-step rollouts using the known pose, velocities and control inputs at step $k$, while the ground-truth velocities $\gtveltwist$ are computed by numerically differentiating the \gls{mocap} pose measurements $\gtpose$.  The \gls{mocap} data was first manually cleaned, to remove any tracking errors. To preserve the generality of the representation, we use body-frame $\veltwist_k$ for learning.

Given that the elements of $\veltwist_k$ corresponding to the actuated DoF of SAM will generally be of larger magnitude than the rest, we first normalize (z-score) the velocities per feature. We similarly normalize $\actinput_k$ per-feature over $\dataset$. All velocities, both in sim and real, are normalized jointly so all velocities share the same distribution. 

We further compute the element-wise mean and standard deviation of $\residual$ across $\dataset$: $\boldsymbol\mu_\delta=\mathbb{E}[\residual]$, \; $\sigma_\delta=\sqrt{\mathbb{V}[\residual]}$, to create the supervised, z-scored residual target:
\begin{equation*}
\tilde{\boldsymbol{\delta}}_k \;=\; \frac{\residual - \boldsymbol\mu_\delta}{\sigma_\delta}
\end{equation*}

\subsection{RESIDUAL DYNAMICS REGRESSION}
We apply a \gls{mlp} architecture to learn the single-step target $\residual$. Our \gls{nn} $\pi_\theta$, depicted in \cref{fig:sim2real_resnet}, consists of 4 blocks, each containing a linear layer with 256 neurons and SiLU activation functions, applied to all the layers except the last one. 
At every timestep, the input layer in $\pi_\theta$ takes in the vector $\boldsymbol{z}_k = \left[ \gtveltwist_{k-1}, \simveltwist_{k}, \actinput_k \right]$ and outputs the inferred residual $\hat{\boldsymbol{\delta}}_k$ such as 
\begin{equation*}
 \hat{\boldsymbol{\delta}}_k = \pi_\theta(\mathbf{z}_k)
\end{equation*}

\subsection{TRAINING}
To minimize the effect of any remaining undetected outliers in the data, during training the target function is a Huber loss \cite{huber1992robust} with $\beta = 0.9$:

\begin{align}
\boldsymbol{r}_k& = \pi_\theta(\mathbf{z}_k) - \tilde{\residual}_k\\
\mathcal{L}_{\beta}(r_k)& =
\begin{cases}
\norm{\boldsymbol{r}_k}^2_2, & \text{if } \norm{\boldsymbol{r}} \le \beta, \\[6pt]
\beta * (\norm{\boldsymbol{r}_k} - \frac{\beta}{2}), & \text{if } \norm{\boldsymbol{r}} > \beta
\end{cases}
\label{eq:sim2real_loss}
\end{align}

We minimize \cref{eq:sim2real_loss} with AdamW \cite{loshchilov2017decoupled} and minibatches $\mathcal{B} = \{r_{n} \}_{n=0}^{N}$ of size $N=768$, applying a weight decay of $10^{-5}$, a learning rate of $3e^{-3}$, and a exponential scheduler with $\gamma = 0.997$ and epochs $=2000$, as follows:

\begin{equation}
    \hat{\theta} = \arg\min_{\theta} \dfrac{1}{N} \sum_{n}^{N} \mathcal{L}_{\beta} (r_{n})
\end{equation}

The training pipeline is depicted in \cref{fig:sim2real_resnet}, on the left side. 

\subsection{INFERENCE}
The inference pipeline is depicted in \cref{fig:sim2real_resnet}, on the right side. 
At inference time, the \gls{nn} employs the simulated velocity  $\veltwist_{k}^{sim}$ instead of $\veltwist_{k}^{gt}$. $\veltwist^{\text{sim}}_{k+1}$ is computed by adding $\hat{\residual}_k$ to the physics in the simulator (PhysX). 
The \gls{nn} output is then mapped back to the un-normalized velocity residual (physical units).
After computing $\veltwist_{k}^{\text{sim}}$ as described in \cref{eq:nom_model}, the corrected simulated velocity during inference is computed as follows:

\begin{equation}
\hat{ \veltwist }_{k}^{\text{sim}} = \veltwist_{k}^{\text{sim}} + \hat{\residual}_k 
\label{eq:sim2real_final_vel}
\end{equation}

\subsection{RESULTS AND DISCUSSION}
The dataset used for initial experiments consists of 8 different recordings of SAM moving through the tank, separated into 24 different segments, based on cuts required for removing \gls{mocap} errors. We used the simulator to resample the data at a constant timestep of 0.02 (50hz), producing a total of 73294 ground truth timesteps equivalent to approximately 24 minutes of real-time operation. To build the $\dataset$ we picked the latest value from each corresponding datastream. Note that the \gls{mocap} data rate\footnote{The \gls{mocap} Data rate is approximately 100 Hz in most rosbags. In practice it varies around 90-110Hz, depending on error rate.} exceeds that of the simulator's, so we did not produce repeated datapoints between successive timesteps.  Out of the 73294 steps, $K = 65096$ were used for training, while two segments totaling 8198 timesteps were used for evaluation. 

The evaluation of the resulting corrected velocities follows the methodology outlined in \cref{sec:sysid-eval}, with $H \in \{1,10,100, 500\}$ and $N=7998$ trajectories from the dataset. 

\begin{table}[h]
\centering
\caption{Evaluation of the residual (res) vs the original (sim) models over 7998 trajectories. RMS of rollout trajectory endpoint errors over time horizon H. Note that a lower number of 7198 trajectories was used for H = 500.}
\label{tbl:sim2real_statistics}
\begin{tabular}{|l|cc|cc|cc|cc|}
\hline
H & $\hat{\pos}^\text{sim}$ & $\pos^\text{sim}$ & $\hat{\eulang}^\text{sim}$ & $\eulang^\text{sim}$  & $\hat{\vel}^\text{sim}$ & $\vel^\text{sim}$ & $\hat{\angvel}^\text{sim}$ & $\angvel^\text{sim}$ \\
\hline
1 & $ 0.00 $ & $ 0.05 $ & $ 0.00 $ & $ 0.14 $ & $ 0.02 $ & $ 0.06 $ & $ 0.08 $ & $ 0.09 $ \\ 
10 & $ 0.02 $ & $ 0.05 $ & $ 0.38 $ & $ 0.42 $ & $ 0.07 $ & $ 0.12 $ & $ 0.19 $ & $ 0.20 $ \\ 
100 & $ 0.20 $ & $ 0.30 $ & $ 0.85 $ & $ 1.20 $ & $ 0.20 $ & $ 0.27 $ & $ 0.23 $ & $ 0.25 $ \\ 
500 & $ 0.32 $ & $ 0.72 $ & $ 0.87 $ & $ 1.20 $ & $ 0.17 $ & $ 0.32 $ & $ 0.25 $ & $ 0.27 $ \\ 
\hline
\end{tabular}
\end{table}

\Cref{tbl:sim2real_statistics} shows the \gls{rmse} of the endpoint pose and velocity components of the overall state error at time horizon $k+H$, for both the normal simulation model $\epsilon^{sim}_{k+H} = \vect{x}^{gt}_{k+H} -\vect{x}^{sim}_{k+H}$ and residual corrected physics model $\hat{\epsilon}^{sim}_{k+H} = \vect{x}^{gt}_{k+H} -\hat{\vect{x}}^{sim}_{k+H}$. The endpoint is determined by initiating from MoCap ground-truth $\vect{x}^{gt}_{k}$ and rolling out up to the residual-corrected $\hat{\vect{x}}^{sim}_{k+H}$ and un-corrected $\vect{x}^{sim}_{k+H}$ vehicles states, which are computed integrating \cref{eq:sim2real_final_vel}. 

\begin{figure}[h]
    \centering

    \subfloat[Example evaluation trajectory with an effective residual correction. Reversing with a slow forward acceleration at the end.]{%
        \includegraphics[width=0.99\linewidth]{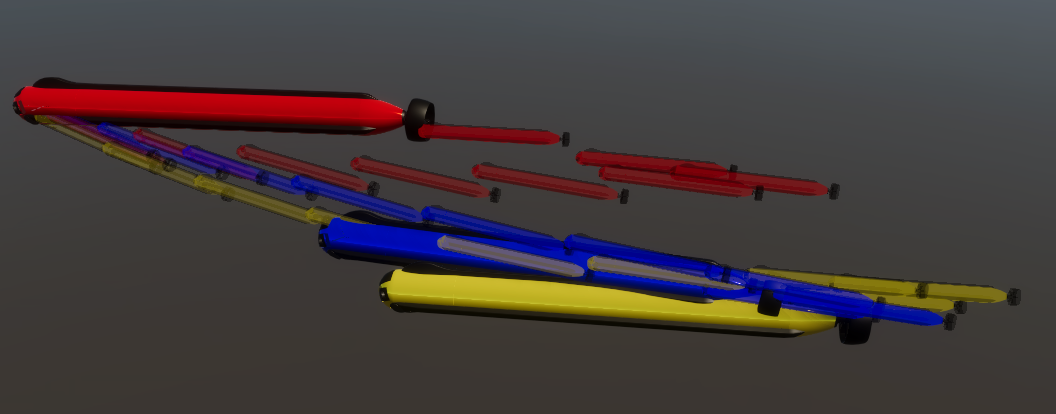}%
    }\\[0.5em] 

    \subfloat[Example evaluation trajectory with unstable residual correction. Buoyancy change accompanied by fast forward acceleration.]{%
        \includegraphics[width=0.99\linewidth]{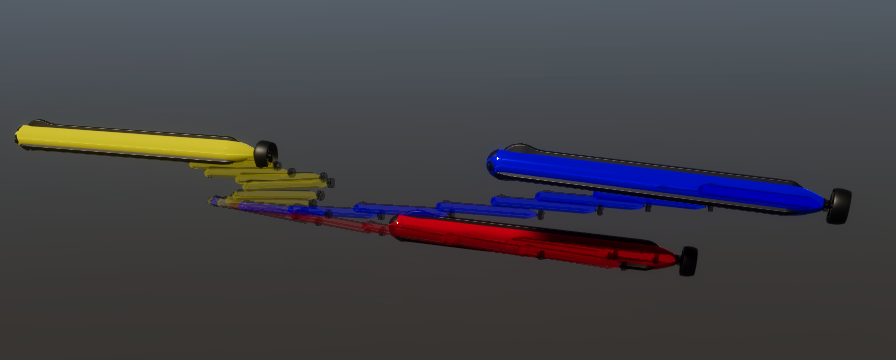}%
    }

    \caption{Two extreme examples of residual model evaluation trajectories in the simulator. The ground truth (yellow), original simulation (red) and corrected simulation (blue) trajectories start on the same pose and receive the same control inputs sequence. Intermediate trajectory models scaled down for clarity.}
    \label{fig:sim2real_results}
\end{figure}

From the quantitative results above it can be seen that the poses and linear velocity errors show overall improvements for longer time horizons. However, we note that there is no significant reduction in angular velocity error. Furthermore, qualitative observations show that when the motion includes significant turns with high angular velocity and sharp changes in forward momentum, the model can become unstable and accelerate in an uncontrollable manner. To illustrate this, we provide two trajectories in \cref{fig:sim2real_results} - one good and one with a relatively poor performance.

Overall, our results highlight the feasibility and potential of applying real-to-sim techniques through the combination of the Marinarium and the SMaRCSim to produce more realistic underwater simulators.
The simple methodology presented to exemplify this already offers considerable improvements to the raw simulation model. However, for real-world deployment, robustness would need to be improved and model instabilities addressed. One idea to achieve a more robust model could be to separate the linear and angular components and model their residuals separately. Alternatives also include recursive neural network architectures to improve the handling of rapid velocity and actuation changes, improving data coverage of possible state-action pairs or different principles for the training process, such as those inspired by reinforcement learning \cite{Johannik2019} or DAgger (Dataset Aggregation) \cite{Ross2010ARO}. 

\section{RESEARCH AREA 4: UNDERWATER VALIDATION OF SPACECRAFT AUTONOMY}
\label{sec:space}
Building on the parallels between neutrally buoyant vehicles and free-flying spacecraft, we harness the connection between the Marinarium and the \gls{kth} space robotics laboratory~\cite{roque2025towards} to investigate how underwater platforms can be used to validate spacecraft guidance and control algorithms. 
Neutral buoyancy enables full 6-DoF actuation and weightless behavior, similar in spirit to the neutral-buoyancy environments used for astronaut training, while hydrodynamic effects introduce the main deviations from true microgravity. To give an initial assessment of how well a space control architecture transfers across domains, we conduct paired experiments in which an underwater ROV and a planar space-robot emulator execute the same autonomous inspection task under an identical software and control stack as shown in \cref{fig:atmos_brov}.

\begin{figure*}[ht!]
    \centering
    \includegraphics[width=0.98\textwidth]{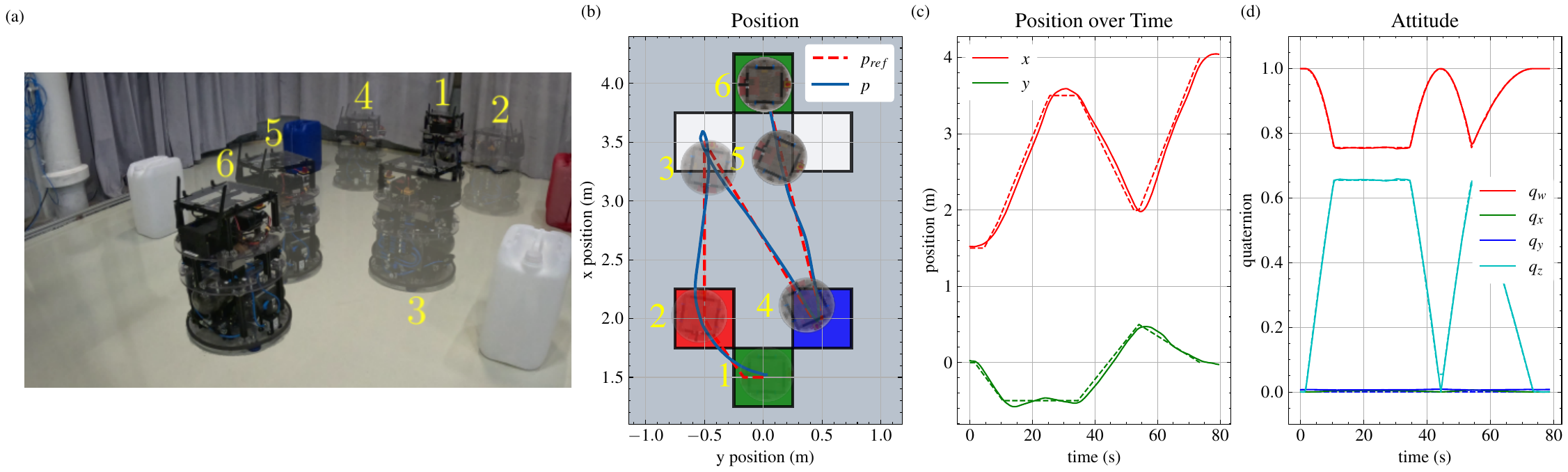}
    \caption{Result from the spacecraft-autonomy experiment using the \gls{atmos} free-flyer. The robot is tasked with inspecting a blue object, a red object, and one of two white objects, followed by moving to a final pose, captured in an STL specification. The motion plan is restricted to a 2D plane. a) Time-lapse from the experiment. b) Top-view trajectory of the executed path. c) Position over time. d) Orientation over time.}
    \label{fig:2D-atmos}
\end{figure*}

\begin{figure*}[t!]
    \centering
    \includegraphics[width=0.98\textwidth]{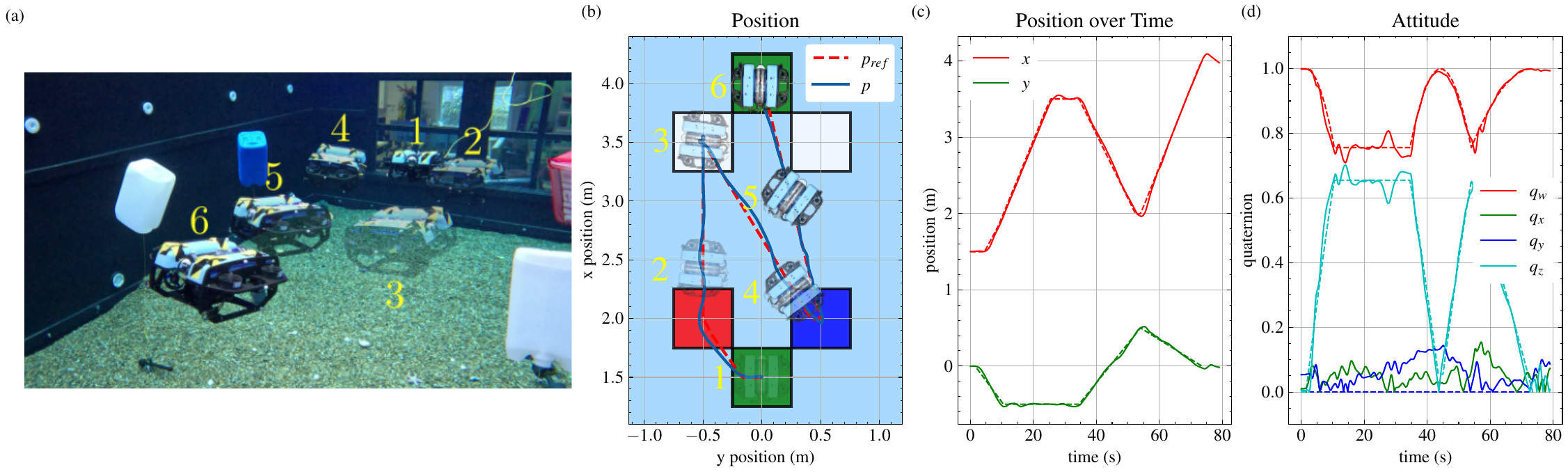}
    \caption{Results from the same spacecraft-autonomy task shown in \cref{fig:2D-atmos}, but executed with a BlueROV underwater robot using the same controller and STL-based plan. Although the plan is two-dimensional, the experimental data reveal some motion in all six dimensions.}
    \label{fig:2D-bluerov}
\end{figure*}

\subsection{EXPERIMENTAL PLATFORMS AND CONTROL ARCHITECTURE}
As introduced in \cref{sec:facility}, a set of BlueROV2 Heavy configuration underwater vehicles have been modified to act as underwater surrogates of \gls{atmos} platforms. The \gls{atmos} vehicle floats on air bearings over an epoxy-coated floor, providing nearly frictionless planar 3-DoF motion control through eight compressed-air thrusters. The BlueROV2 Heavy is equipped with eight bidirectional propellers that enable full 6-DoF motion underwater. For comparability, only planar trajectories are commanded during this experiment to match the \gls{atmos} configuration.

\subsection{CONTROLLER FORMULATION}
Both systems are controlled using the same 6-DoF free-flyer dynamics model. Translational motion is expressed in the inertial frame through the position $\pos$, while the linear velocity $\vel$ is expressed in the body frame. Attitude is parameterized by a unit quaternion $\quat$ describing the rotation from the body frame to the inertial frame, with the corresponding rotation matrix denoted $R(\quat)$. The state is then defined as 
\[
\vect{x} \doteq [\pose^\top,\veltwist^\top]^\top \doteq [\pos^\top,\quat^\top,\vel^\top,\angvel^\top]^\top \in \mathbb{R}^{13},
\]
and the control input is defined as $\wrench = [\actforce^\top,\vect{\tau}^\top]^\top$ with $\actforce$ and $\vect{\tau}$ the commanded body frame force and torque.
The Newton-Euler equations of motion are
\begin{subequations}
\label{eq:ff_model}
\begin{align}
\dot{\pos} &= R(\quat) \vel \\
\dot{\quat} &= \tfrac{1}{2}\Omega(\angvel) \quat \\
\dot{\vel} &= \tfrac{1}{m} \left(\actforce + R(\quat)^{T} \distforce \right) \\
\dot{\angvel} &= J^{-1}\left(\vect{\tau} + \vect{\tau}_d - \angvel \times J\angvel \right)
\end{align}
\end{subequations}
where $m$ is the vehicle mass, $J$ is the inertia matrix, and $\angvel$ is the body-frame angular velocity. External disturbances and model mismatches (including hydrodynamic effects) are represented by the forces $\distforce$ and torques $\vect{\tau}_d$. The operator $\Omega(\angvel)$ denotes the quaternion kinematic matrix built from the angular velocity. This model captures the essential inertial, torque-driven motion of a free-flying spacecraft while allowing both platforms to be controlled using an identical NMPC formulation.

High-level force and torque commands are computed by solving a nonlinear MPC problem formulated as the following optimal control problem (OCP):
\begin{subequations}
\label{eq:nmpc_ocp}
\begin{align}
J^*(x_k) &= \min_{\wrench_k} \sum_{n=0}^{N-1} \ell(\vect{x}(n|k), \wrench(n|k)) + \ell_f(\vect{x}(N|k)) \\
\text{s.t.} \quad & \vect{x}(0|k) = \vect{x}_k \\
& \vect{x}(n+1|k) = f\big(\vect{x}(n|k), \wrench(n|k)\big) \\
& \vect{x}(n|k) \in \mathbb{X}, \quad \wrench(n|k) \in \mathbb{U}.
\end{align}
\end{subequations}
The nonlinear dynamics function $f(\cdot)$ is obtained by discretizing the continuous-time model \cref{eq:ff_model} with a sampling time $T_s$.
The sets $\mathbb{X}$ and $\mathbb{U}$ encode state and input constraints such as workspace limits and actuator bounds. The stage and terminal costs $\ell(\cdot)$ and $\ell_f(\cdot)$ penalize deviations from a reference trajectory
\begin{align}
\ell\big(\vect{x}, \wrench\big)
&= \big\|\vect{x} - \vect{x}_r\big\|_{Q}^2
 + \big\|\wrench - \wrench_r\big\|_{R}^2, \\
\ell_f\big(\vect{x}\big)
&= \big\|\vect{x} - \vect{x}_r\big\|_{P}^2,
\end{align}
with positive semi-definite weighting matrices $Q, R, P$.

The OCP is solved in real time using ACADOS~\cite{Verschueren2021}, and the resulting control inputs are transmitted to the micro-controller which allocates the actuators such that the desired force and torque is achieved on either space- or underwater platform. An \gls{ekf} estimates external disturbance forces and torques during operation, which primarily affect the ROV due to unmodeled hydrodynamic effects, parameter uncertainties, and tether disturbances. The shared control layout is shown in \cref{fig:atmos_brov}.

Both platforms execute a planned trajectory that satisfies a complex inspection task formulated as a Signal Temporal Logic specification~\cite{donze2010robust}. The specification requires sequential inspection of the blue, the red, and either of the white objects (with some temporal flexibility) in a planar workspace. The plan is defined in two dimensions to allow direct comparison of the systems' behavior.

\subsection{RESULTS AND DISCUSSION}
\begin{figure}[ht!]
    \centering
    \includegraphics[width=0.95\linewidth]{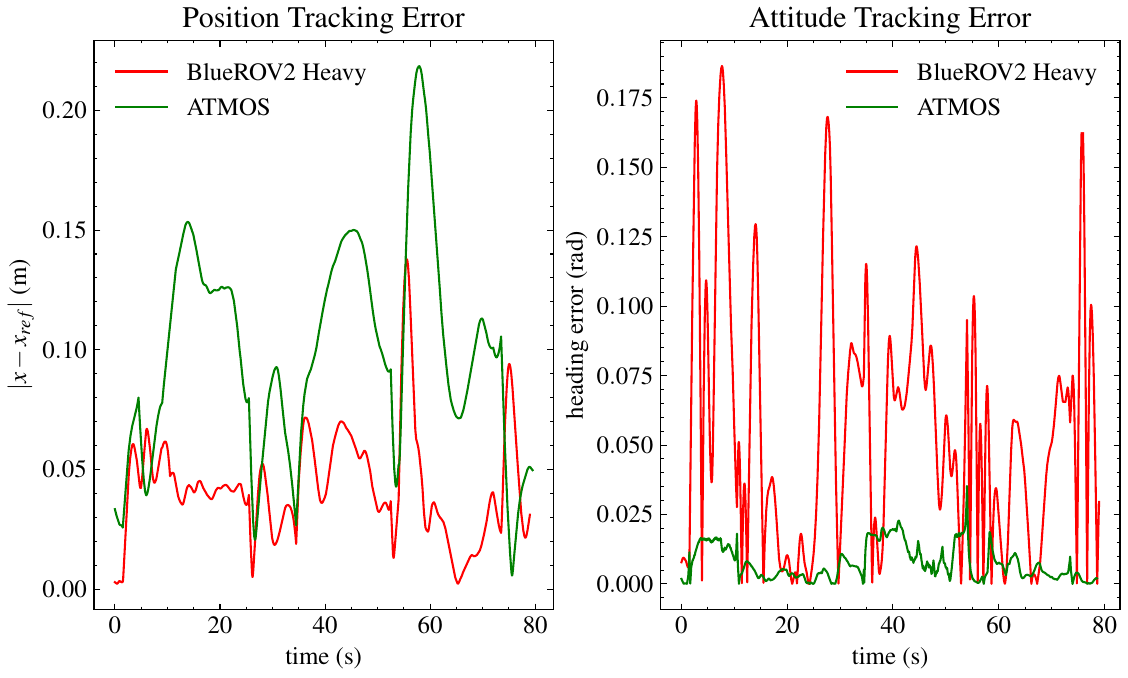}
    \caption{Tracking errors of the space validation experiment on the \gls{atmos} free-flyer and the BlueROV.}
    \label{fig:atmos_bluerov_tracking_errors}
\end{figure}

The resulting position and attitude trajectories are shown in \cref{fig:2D-atmos} and \cref{fig:2D-bluerov}. Both platforms successfully track the STL-generated plan and exhibit similar overall motion profiles, demonstrating that the underwater system can reproduce the behavior of a free-flying spacecraft under the same control architecture. The tracking errors w.r.t. the reference trajectory are highlighted in \cref{fig:atmos_bluerov_tracking_errors}.
The BlueROV2 displays small oscillations in attitude, which are expected due to unmodeled hydrodynamic forces and residual tether disturbances. Nevertheless, its trajectory remains consistent with the reference, validating the feasibility of underwater testing for spacecraft guidance and control methods.

These results illustrate the value of the Marinarium's tight integration with the KTH space robotics laboratory, demonstrating how the facility can be used to validate spacecraft guidance and control algorithms under realistic dynamic conditions. Extending this approach to six degrees of freedom, including active compensation of known hydrodynamic effects, and establishing formal guarantees for equivalence between underwater and space-domain validation, represent important research directions for these facilities.

\section{CONCLUSION}
This work has introduced the Marinarium, a modular, cost-effective underwater robotics research facility conceived as an intermediate field-robotics testbed for marine and space-analog autonomy.
By combining dense dual-domain \gls{mocap}, a retractable roof, a SMaRCSim digital twin, and integration with a space-robotics infrastructure, the facility narrows the gap between desktop simulation and costly offshore or on-orbit campaigns while remaining accessible to academic groups.

\paragraph*{Lessons for field robotics.}
Our experience suggests three principles for field robotics development.

(1)~\emph{Observable cross-domain integration reduces field time}: tank testing with MoCap ground truth enables a consistent development of capabilities that is impractical in open water. While limited to controlled conditions, these steps reduce the risk of costly field deployments and accelerate the development process.

(2)~\emph{Coupled physical and digital twins reduce transfer risk}: real-to-sim residual learning measurably improves dynamics fidelity, although currents, large-scale disturbances, and perception effects remain for open-water study.

(3)~\emph{Shared autonomy stacks enable cross-domain validation}: deploying an identical software pipeline across space and underwater surrogates makes it possible to compare mission-level behavior across domains before committing to on-orbit or offshore deployment.

Furthermore, the four case studies in this article yield the following technical contributions:
\begin{itemize}
    \item Koopman \gls{edmdc} identification improves long-horizon prediction of BlueROV2 dynamics relative to physics-based and learning baselines, highlighting both the value of instrumented tanks and the need for further \gls{uw} modeling research.
    \item A confined heterogeneous air-surface-underwater rendezvous mission demonstrates how multi-domain fleets can be stress-tested under controlled conditions before offshore deployment.
    \item Learned dynamics residuals, regressed from paired tank and simulator data, reduce open-loop simulation error, illustrating a practical Sim2Real pathway for underwater platforms.
    \item Underwater surrogates enable direct comparison of missions on planar space and neutrally buoyant platforms, supporting underwater validation of spacecraft guidance and control.
\end{itemize}
Taken together, these results show how deliberately coupling precise sensing, cross-domain access, and simulation integration can accelerate the path from laboratory development to field-ready maritime and space robotics. Future work will migrate algorithms validated in the Marinarium to open-water trials within the SMaRC test ecosystem.


\bibliography{references}

\vfill\pagebreak

\end{document}